\documentclass[10pt,twocolumn,letterpaper]{article}
            
\usepackage{cvpr}      

\usepackage{booktabs}                     
\usepackage{multirow}                 
\usepackage{multicol}                 
\usepackage{multirow}                
\usepackage{float}                    
\usepackage{makecell}                 
\usepackage{booktabs}                 
\usepackage{newfloat}
\usepackage{listings}
\usepackage{diagbox}
\usepackage{comment}
\usepackage{bm}

\usepackage{amsmath}
\usepackage{color}
\usepackage{colortbl}
\usepackage{xcolor}
\usepackage[abs]{overpic}
\usepackage{algpseudocode}
\usepackage[linesnumbered,ruled,vlined]{algorithm2e}
\usepackage[accsupp]{axessibility}

\definecolor{cvprblue}{rgb}{0.21,0.49,0.74}
\usepackage[pagebackref,breaklinks,colorlinks,allcolors=cvprblue]{hyperref}

\def\paperID{4325} 
\def\confName{CVPR}
\def\confYear{2025}

\newcommand*{\affaddr}[1]{#1} 
\newcommand*{\affmark}[1][*]{\textsuperscript{#1}}

\title{Learning Compatible Multi-Prize Subnetworks for Asymmetric Retrieval}

\author{Yushuai Sun\affmark[1*], Zikun Zhou\affmark[2*$\dagger$], Dongmei Jiang\affmark[2], Yaowei Wang\affmark[2],\\Jun Yu\affmark[1], Guangming Lu\affmark[1], and Wenjie Pei\affmark[1,2$\dagger$]\\\affaddr{\affmark[1]Harbin Institute of Technology, Shenzhen}\quad\affaddr{\affmark[2]Pengcheng Laboratory}\\
}

\begin{document}
\maketitle
\renewcommand{\thefootnote}{\fnsymbol{footnote}} 
\footnotetext{*These authors contribute equally.}
\footnotetext{$\dagger$Zikun Zhou and Wenjie Pei are corresponding authors (zhouzikunhit@gmail.com, wenjiecoder@outlook.com).}
\begin{abstract}
Asymmetric retrieval is a typical scenario in real-world retrieval systems, where compatible models of varying capacities are deployed on platforms with different resource configurations. Existing methods generally train pre-defined networks or subnetworks with capacities specifically designed for pre-determined platforms, using compatible learning. Nevertheless, these methods suffer from limited flexibility for multi-platform deployment. For example, when introducing a new platform into the retrieval systems, developers have to train an additional model at an appropriate capacity that is compatible with existing models via backward-compatible learning. In this paper, we propose a Prunable Network with self-compatibility, which allows developers to generate compatible subnetworks at any desired capacity through post-training pruning. Thus it allows the creation of a sparse subnetwork matching the resources of the new platform without additional training. Specifically, we optimize both the architecture and weight of subnetworks at different capacities within a dense network in compatible learning. We also design a conflict-aware gradient integration scheme to handle the gradient conflicts between the dense network and subnetworks during compatible learning. Extensive experiments on diverse benchmarks and visual backbones demonstrate the effectiveness of our method. Our code and model are available at \url{https://github.com/Bunny-Black/PrunNet}.
\end{abstract}    

\section{Introduction}
\label{sec:intro}

\begin{figure}[t!]
\centering
\includegraphics[width=1.0\linewidth]{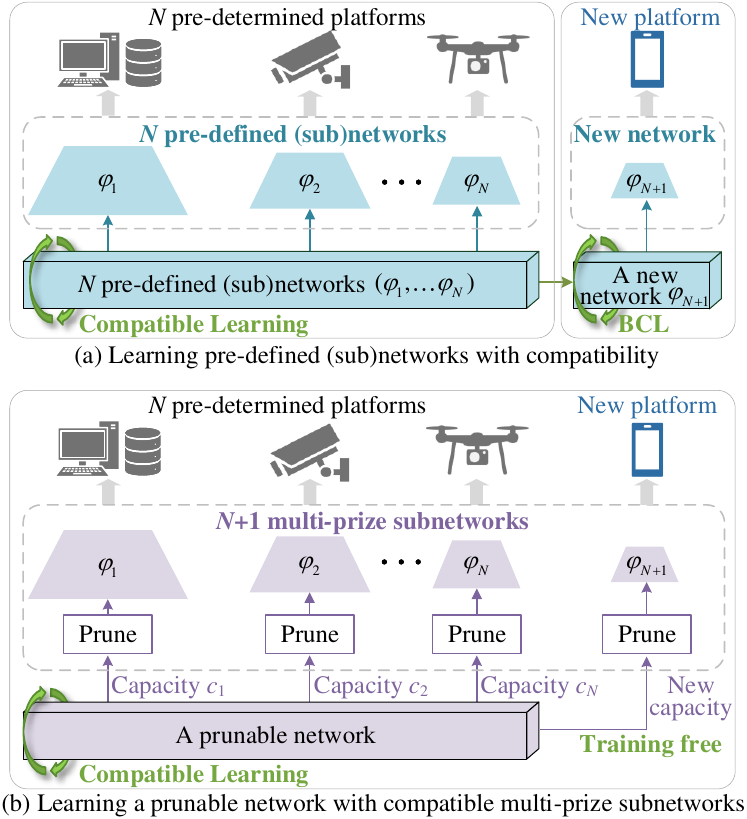}
\caption{Two pipelines for learning compatible models with different capacities for multi-platform deployment. (a) Existing methods tailor $N$ pre-defined (sub)networks for pre-determined platforms through compatible learning and train additional models for new platforms by Backward-Compatible Learning (BCL). (b) Our method constructs a prunable network that can generate compatible subnetworks at any specified capacity via pruning.}
\label{Fig:introduction}
\end{figure}

Image retrieval~\cite{ImageRetrieval2008,ImageRetrieval2023,ImageRetrieval2024} has been extensively studied for many years. Traditional retrieval systems use the same model to process both query and gallery images, known as \emph{symmetric retrieval}. Nevertheless, symmetric retrieval is not always optimal for real-world applications involving devices with diverse computation and storage resources, such as cloud servers and mobiles. Deploying a lightweight model tailored for the device with minimal resources would limit the performance and waste the resources of the other devices. To address the issue, many studies~\cite{wu2023_SFSC,xie2023_CDD,xie2024_d3still,wu2022_CSD} explore asymmetric retrieval, training multiple retrieval models with different capacities and deploying them on different devices. Typically, the large-capacity model is deployed on a cloud server to index gallery images, while the small-capacity one is deployed on a resource-constrained device to process query images. They are referred to as the gallery and query models, respectively.

Asymmetric retrieval requires compatibility between the gallery and query models, meaning that similar images processed by different models are mapped closer in the feature space, while dissimilar images are placed farther apart. Many asymmetric retrieval methods~\cite{wu2022_CSD,xie2024_d3still,suma2023_RAML} resort to knowledge distillation to obtain a lightweight student model that is compatible with the heavyweight teacher model. Besides, several methods~\cite{wu2023_AFF,duggal2021_CMP_NAS} adopt the classifier of the large-capacity gallery model to regularize the small-capacity query model. 

These algorithms mainly focus on learning a single small-capacity model. The recently proposed method, SFSC~\cite{wu2023_SFSC}, aims to simultaneously learn compatible models of different capacities for multi-platform deployment. Specifically, SFSC~\cite{wu2023_SFSC} introduces a switchable network containing several pre-defined subnetworks and optimizes these subnetworks through a compatible loss. Thus, any two subnetworks within SFSC are compatible, a property referred to as ``self-compatibility''.

Figure~\ref{Fig:introduction} (a) summarizes existing methods for acquiring compatible models of different capacities, which train independent networks or parameter-sharing subnetworks with compatible constraints. A limitation is that the architectures of these (sub)networks are pre-defined prior to model training. Given $N$ pre-determined platforms, developers can employ the methods to train $N$ pre-defined (sub)networks tailored to match the resource constraints of the platforms. However, when a new platform is introduced to the retrieval system, these methods cannot directly produce a model with a suitable capacity. Developers have to train an additional network compatible with existing models via Backward-Compatible Learning (BCL). Besides, SFSC uses pre-defined and fixed architectures for the parameter-sharing subnetworks in compatible learning, restricting the optimization space to find the optimal subnetworks.

In this paper, we explore optimizing both the architecture and weight of subnetworks at different capacities within a dense network in compatible learning. Specifically, we aim to discover effective subnetwork architectures, rather than pre-defining and fixing them, inspired by the Lottery Ticket Hypothesis (LTH)~\cite{What's_hidden,Supertickets}. The LTH researches demonstrate the existence of sparse subnetworks, known as ``winning tickets'', within a dense network, which can achieve comparable performance with the dense network. Differently, our goal is to identify well-performing subnetworks at each specified capacity. We refer to these well-performing subnetworks as ``multi-prize subnetworks''.
We begin with preliminary experiments using edge-popup~\cite{What's_hidden} to investigate weight reuse across the well-performing subnetworks at different capacities. The results provide a key insight: \emph{a small-capacity prize subnetwork can be obtained by selectively inheriting weights from a large-capacity prize subnetwork, rather than searching for it within the entire dense network.} It means that we can identify multi-prize subnetworks of various capacities through greedy pruning. 

Based on this observation, we design a Prunable Network (PrunNet) with self-compatibility, which allows developers to generate compatible subnetworks at any desired capacity through post-training pruning, as shown in Figure~\ref{Fig:introduction} (b). It allows the creation of a sparse subnetwork suitable for new platforms without retraining. Specifically, we assign a learnable score for each weight,~\ie, neural connection, of the dense network, which indicates the importance of the weight. We perform greedy pruning on the dense network during optimization. Hence, the architecture of the subnetworks can be optimized along with the updating of the scores.
Besides, we design a conflict-aware gradient integration scheme to solve the gradient conflicts between the (sub)networks during compatible learning. Extensive experiments on diverse benchmarks and visual backbones demonstrate the effectiveness of the proposed method. Our contributions are summarized as follows:
\begin{itemize}
    \item We propose a Prunable Network (PrunNet) which can generate compatible subnetworks at any specified capacity through greedy pruning after model training.
    \item We propose a conflict-aware gradient integration scheme to find an optimization direction in agreement with the majority of the losses, which mitigates the impact of the conflicting gradients during training PrunNet.
    \item Extensive experiments on various benchmarks demonstrate that our method outperforms the existing approaches in both discriminability and compatibility. 
\end{itemize}

\section{Related work}
\noindent\textbf{Compatible learning.}
Compatible Learning aims to generate cross-model comparable features. A typical application is Asymmetric Retrieval~\cite{budnik2021_AML,wu2023_SSPL,shoshan2024_ensemble}, where query models of varying capacities are trained to be compatible with the large gallery model, achieving a trade-off between performance and deployment flexibility. Knowledge distillation~\cite{budnik2021_AML,suma2023_RAML,xie2023_CDD,wu2022_CSD,wu2023_ROP,xie2024_d3still}, is widely used to learn a light-weight query model compatible with a heavy-weight gallery model. Besides, some methods leverage the classifier~\cite{wu2023_AFF,duggal2021_CMP_NAS} of the large-capacity model to regularize the small-capacity one. Neural architecture search is also introduced to train a compatible model~\cite{duggal2021_CMP_NAS}. 
Recently, SFSC~\cite{wu2023_SFSC} is proposed to simultaneously learn compatible models with different capacities for multi-platform deployment. SFSC introduces a Switchable Network (SwitchNet) containing several pre-defined subnetworks and optimizes the subnetworks through a compatible loss.

Compatible learning is also used to train a new model backward-compatible with the old one, upgrading the retrieval model without backfilling~\cite{BCT}. BCT~\cite{BCT} achieves backward compatibility by aligning the new model to the old one in the logit space,~\ie, regularizing the new model using the classifier of the old one. The other methods explore sophisticated compatible constraints, such as contrastive loss~\cite{hot-refresh,Online-Backfilling,dual-tuning,Prototype_Perturbation} and boundary loss~\cite{advbct,LCE}. Unlike most existing methods using pre-defined architectures, we learn both the architecture and weight of the subnetworks at various capacities within a dense network in a compatible learning manner for multi-platform deployment.

\vspace{1mm}
\noindent\textbf{Lottery ticket hypothesis.} The Lottery Ticket Hypothesis (LTH)~\cite{LTH} states that a dense network contains sparse subnetworks (\ie, winning tickets) that can achieve comparable performance to the original network in a similar number of iterations. Subsequent works~\cite{What's_hidden,MPT} use an edge-popup algorithm to find subnetworks within a randomly initialized network that can achieve good performance without training. Edge-popup~\cite{What's_hidden} optimizes all scores to find a good subnetwork within the dense network while keeping the weight frozen.
Additionally, some methods combine pruning with weight optimization to progressively identify winning ticket sub-models, as exemplified by SuperTickets~\cite{Supertickets}, which prunes at fixed intervals during training. LTH has also been applied in incremental learning. WSN~\cite{WSN} learns a winning subnetwork for the novel task while keeping the weights of previous tasks frozen to mitigate catastrophic forgetting. Differently, our method aims to find well-performing and compatible subnetworks at various specified capacities, and we optimize both the scores and weights to learn hierarchically pruned subnetworks.

\vspace{1mm}
\noindent\textbf{Multi-task learning.}
Multi-Task Learning (MTL)~\cite{multi-task-learning} is a paradigm learning multiple related tasks jointly, leveraging the shared knowledge to improve the generalization for individual tasks. A primary challenge in MTL is conflicting gradients, where gradients for different tasks diverge significantly, potentially hindering model convergence and resulting in poor generalization~\cite{PCGrad, MGDA}. To address this issue, several methods~\cite{MGDA, CAGrad, MTL_uncertainty, Gradnorm} resort to Pareto optimization, which resolves conflicts by learning task-specific gradient weighting coefficients. Additionally, some approaches mitigate conflicts by directly modifying the gradients~\cite{PCGrad, Grad_Vaccine}. For instance, PCGrad~\cite{PCGrad} projects the gradient vector of one task onto the normal plane of its conflicting counterparts. Unlike the methods, we propose a conflict-aware gradient integration method to alleviate conflicts.

\section{Methodology}
In this section, we introduce PrunNet, a network capable of generating compatible subnetworks with any specified capacity. We begin by presenting insights into the key characteristics of prize subnetworks and subsequently present the design of PrunNet and the details of model optimization.

\subsection{Weight inheritance in multi-prize subnetworks}
Our goal is to discover and optimize multiple well-performing subnetworks at various capacities within a dense network, \ie, multi-prize subnetworks. To this end, we begin with preliminary experiments to investigate weight reuse between two identified prize subnetworks, which inform the design of our method. Herein we employ the edge-popup algorithm~\cite{What's_hidden}, which learns a set of capacity-conditioned scores to identify a good subnetwork from a randomly initialized network. Besides One-Shot Pruning (OSP) proposed in~\cite{What's_hidden}, we also perform Iterative Pruning (IP) using edge-popup. OSP optimizes the capacity-conditioned scores in a single round to directly identify a good subnetwork, while IP progressively prunes the dense network step-by-step by learning the scores conditioned on a capacity factor that decreases stepwise. In each step, IP attempts to identify a small, good subnetwork from the larger subnetwork identified in the previous step, rather than directly from the dense network.

\begin{figure}[t]
\centering
\includegraphics[width=0.9\linewidth]{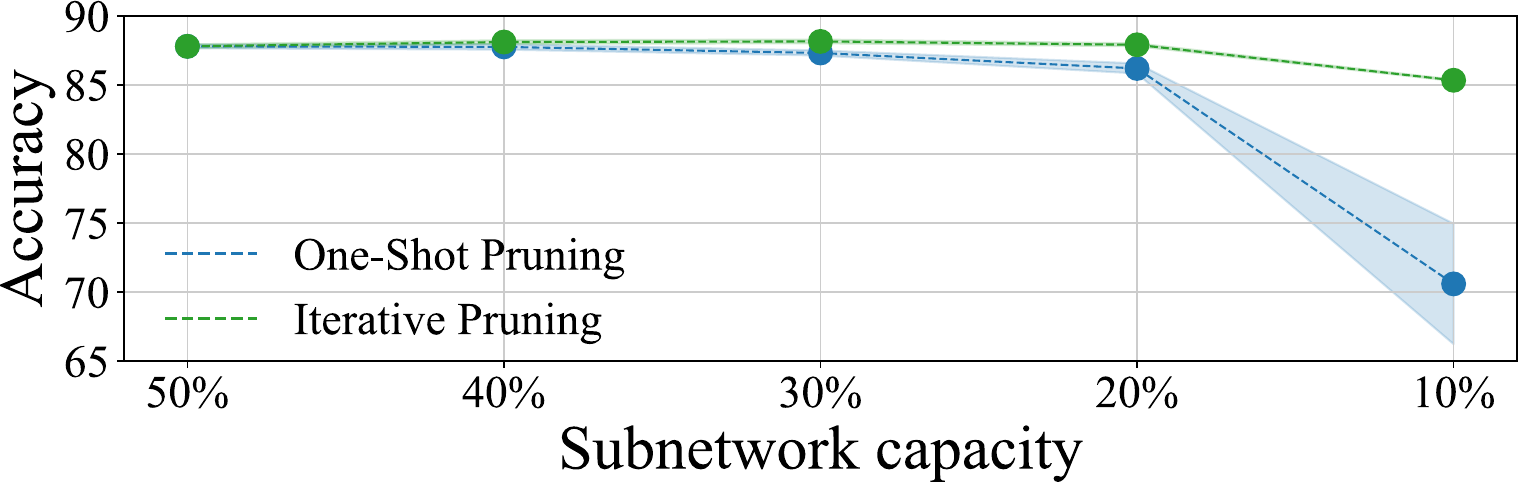}
\vspace{-2mm}
\caption{Comparisons between one-shot and iterative pruning with edge-popup~\cite{What's_hidden}. The plots show the mean results on 5 random initializations. Shading areas denote the standard deviation.}
\label{Fig: submodel overlap}
\vspace{-2mm}
\end{figure}

Figure~\ref{Fig: submodel overlap} presents the classification accuracy on CIFAR-10~\cite{CIFAR} of the identified subnetworks. Empirically, we obtain a crucial insight into the multi-prize subnetworks: \emph{A small prize network found from a large prize network is also the prize subnetwork of the dense network, evidenced by the superior accuracy of IP compared with OSP. Thus we can obtain a small-capacity prize subnetwork by selectively inheriting weights from a large-capacity prize subnetwork, rather than searching for it within the entire dense network.} Please refer to \textbf{Appendix A} for more details and analyses.

\begin{figure*}[t!]
\centering
\includegraphics[width=0.9\linewidth]{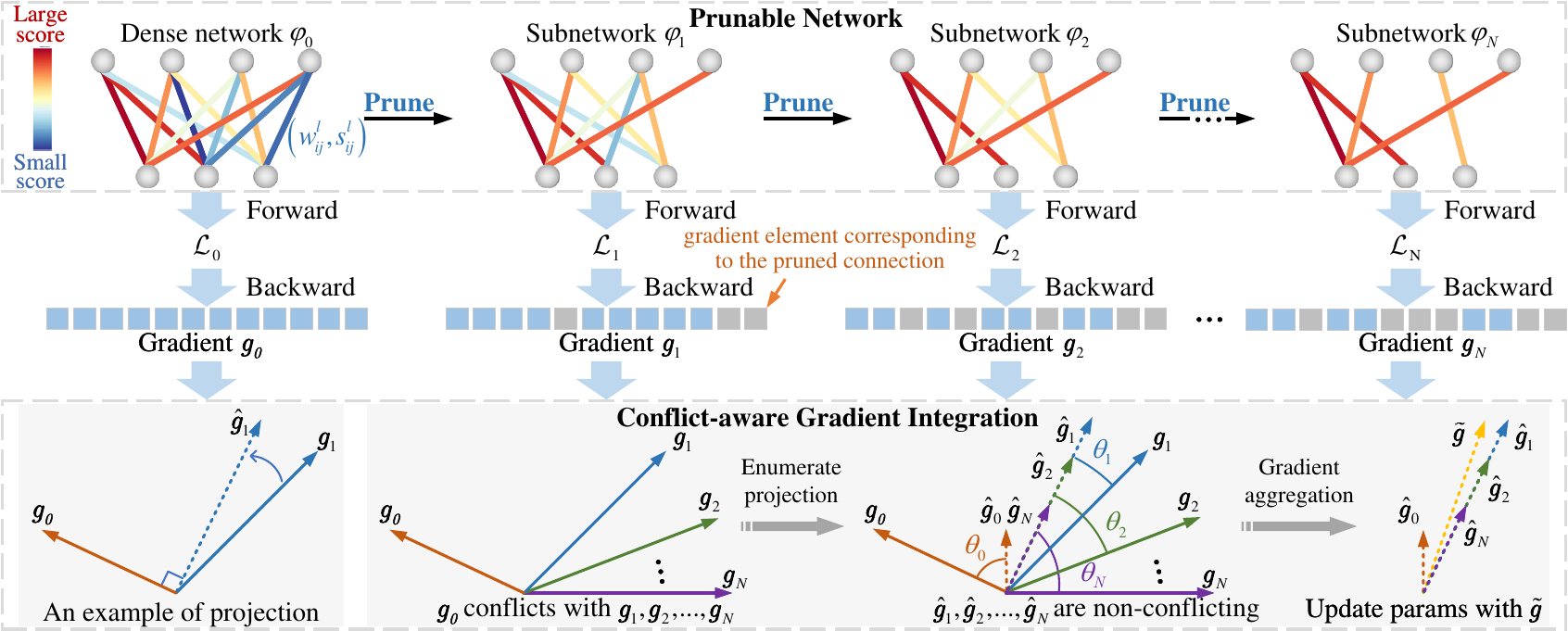}
\vspace{-1mm}
\caption{Overall pipeline for constructing and optimizing a Prunable Network (PrunNet). Each connection in PrunNet is characterized by a weight $w^l_{ij}$ and a score $s^{l}_{ij}$. The subnetworks are generated by greedy pruning according to the scores. After calculating the gradient of the losses $\{\mathcal{L}_0, \mathcal{L}_1,...,\mathcal{L}_N\}$, we use conflict-aware gradient integration to obtain the gradient $\tilde{\bm g}$ updating the parameters of PrunNet.}
\label{Fig:Method}
\vspace{-0.5mm}
\end{figure*}

\subsection{Prunable network}
Inspired by the weight inheritance nature in multi-prize subnetworks, we propose a Prunable Network (PrunNet) with self-compatibility, enabling developers to derive compatible prize subnetworks at arbitrary capacities through greedy pruning, as illustrated in Figure~\ref{Fig:Method}. Specifically, we assign a learnable score to each weight of the traditional dense network $\phi_0$, which is updated alongside the weight during optimization. The connection between the two neurons is characterized by both a weight and a score, representing the strength and the importance of the connection, respectively. With the score map, we can adopt a greedy connection-pruning strategy to remove less important connections, generating sparse subnetworks of various capacities. In this way, small subnetworks inherit the connections from larger subnetworks. Technically, to obtain a subnetwork $\phi_i$ with a capacity factor of $c_i\%$, we retain only the connections with the top-$c_i\%$ scores and discard the others.

Considering that the resource limitation is more related to the width of layers than the number of layers, we reduce the dense network width to $c_i\%$ of its original width by pruning, following SFSC~\cite{wu2023_SFSC}. Specifically, we retain $c_i\%$ of the connections in each layer. Taking a fully connected network as an example, the input $\mathcal{I}^{l}_i$ to the $i$-th neuron at the $l$-th layer $n^{l}_i$ can be formulated as:
\begin{equation}
\label{eq:sub_forward}
\setlength{\abovedisplayskip}{3pt}
\setlength{\belowdisplayskip}{3pt}
    \mathcal{I}^{l}_i = \sum_{j=1}^{M^{l-1}}r(s^{l}_{ij})w^{l}_{ij}\mathcal{Z}^{l-1}_j,
\end{equation}
where $w^{l}_{ij}$ and $s^{l}_{ij}$ are the weight and score of the connection between $n^{l}_i$ and $n^{l-1}_j$, respectively. $M^{l-1}$ is the neuron number in the previous layer. $r(s^{l}_{ij})\!=\!1$ if $s^{l}_{ij}$ belongs to the top-$c_i\%$ scores in the $l$-th layer, and $r(s^{l}_{ij})\!=\!0$ otherwise. $\mathcal{Z}^{l-1}_j$ is the activated output of $n^{l-1}_j$. Similar pruning operations can also be applied to the convolutional layers. It means our greedy pruning mechanism can be applied to both the convolution and transformer architectures. Note that we do not apply pruning to the normalization layers, which constitute a small proportion of the total parameters. Moreover, the normalization layers are shared across all subnetworks. Although our pruning method is unstructured, the resulting sparse subnetworks can be efficiently accelerated on various hardware platforms~\cite{Sparsert, FPGA_accelerator, NPU_accelerate_2, Flexagon}.

In PrunNet, both the weights and scores are learnable. It means that the parameters and architectures of the subnetworks are optimized during model training. Notably, multiple subnetworks of various capacities are optimized simultaneously, so that the learned scores can accurately rank the connections (\ie, the weights) by importance. Through a single training process, the learned scores enable the selection of the most important connections at any specified proportion to form a prize subnetwork,~\ie, post-training pruning. In contrast, Edge-popup~\cite{What's_hidden} optimizes the scores alone and selects a predefined proportion of connections with randomly initialized weights. Next, we outline the optimization procedure of PrunNet.

\subsection{Compatible learning for prunable network}
The compatible learning process for PrunNet involves several subnetworks with various capacities. Specifically, we pre-define $N$ capacity factors $\{c_i\}_{i=1}^{N}$ and accordingly derive $N$ subnetworks $\{\phi_i\}_{i=1}^{N}$ through the above-mentioned pruning approach during model training, as illustrated in Figure~\ref{Fig:Method}. To enable both the dense network and the subnetworks to acquire strong discriminability and mutual compatibility, we apply a discriminative loss on the dense network and impose a compatibility constraint on each subnetwork. Without loss of generality, the discriminative loss can be implemented using either cross-entropy or contrastive loss, while the compatibility constraint is enforced by aligning each subnetwork with the dense network in either the embedding space~\cite{LCE} or the logit space~\cite{BCT}. We denote the losses applied to $\{\phi_0,\phi_1,...\phi_N\}$ as $\{\mathcal{L}_0,\mathcal{L}_1,...,\mathcal{L}_N\}$.

Nevertheless, optimizing PrunNet with these losses is challenging due to gradient conflicts between different losses~\cite{PCGrad}. Directly minimizing the sum of the losses would cause the optimizer to struggle to make progress or result in one loss dominating the optimization. To address this conflicting issue, we propose a conflict-aware gradient integration method. Specifically, one iteration of PrunNet involves two steps: 1) performing backward propagation of each loss \emph{w.r.t.} the parameters of PrunNet to derive gradient vectors, and 2) integrating the gradient vectors using a conflict-aware approach to obtain an integrated gradient $\tilde{\bm g}$, which is then used to update the parameters of PrunNet.

\vspace{1mm}
\noindent\textbf{Backward propagation of individual subnetwork.}
As shown in Eq.~\eqref{eq:sub_forward}, the forward propagation of the subnetwork involves a nondifferentiable function $r(\cdot)$, whose output depends on the ranking order of the input. To handle this problem, we use the straight-through gradient estimator~\cite{What's_hidden,Straight-through-Estimator}, treating $r(\cdot)$ as the identity function during backward propagation. Both the weight and score are trainable in our PrunNet. For the loss function $\mathcal{L}_i$, the gradient \emph{w.r.t.} $w^{l}_{ij}$ and $s^{l}_{ij}$ can be formulated as:
\begin{equation}
\setlength{\abovedisplayskip}{4pt}
\setlength{\belowdisplayskip}{4pt}
\label{eq:backward}
\begin{split}
    &\frac{\partial \mathcal{L}_i}{\partial w^{l}_{ij}}=\frac{\partial \mathcal{L}_i}{\partial \mathcal{I}^{l}_i}\frac{\partial \mathcal{I}^{l}_i}{\partial w^{l}_{ij}} =\frac{\partial \mathcal{L}_i}{\partial \mathcal{I}^{l}_i}r(s^{l}_{ij})\mathcal{Z}^{l-1}_j,\\
    &\frac{\partial \mathcal{L}_i}{\partial s^{l}_{ij}}=\frac{\partial \mathcal{L}_i}{\partial \mathcal{I}^{l}_i}\frac{\partial \mathcal{I}^{l}_i}{\partial s^{l}_{ij}}=\frac{\partial \mathcal{L}_i}{\partial \mathcal{I}^{l}_i} w^{l}_{ij} \mathcal{Z}^{l-1}_j.
\end{split}
\end{equation}

\noindent\textbf{Gradient integration and parameter update.}
The gradient values corresponding to multiple parameters form a gradient vector. Generally, the gradient vectors computed with different losses point to different directions. Two gradient vectors are conflicting if their cosine similarity is negative. Instead of directly adding these gradients together, we perform conflict-aware gradient integration to calculate an integrated gradient $\tilde{\bm g}$, aiming to alleviate the impact of the gradient conflicts. Denoting the parameters of PrunNet by $\bm{\theta}$ and the gradient vector computed with loss $\mathcal{L}_i$ by $\bm{g}_i$, the parameter update process can be formulated as:
\begin{equation}
\label{eq:update}
\setlength{\abovedisplayskip}{4pt}
\setlength{\belowdisplayskip}{4pt}
\bm \theta \gets \bm \theta - \eta \psi(\bm g_0, \bm g_1,..., \bm g_N).
\end{equation}
Here, $\psi$ refers to the conflict-aware gradient integration operation, and $\eta$ is the learning rate. Next, we detail the proposed conflict-aware gradient integration approach.

\subsection{Conflict-aware gradient integration}
Figure~\ref{Fig:Method} illustrates the conflict-aware gradient integration process, using an example where $\bm g_0$ is conflicting with $\{\bm g_1, \bm g_2,...,\bm g_N\}$. For a pair of conflicting gradients, we first project each of them onto the orthogonal plane of the other to eliminate the conflicting components, inspired by~\cite{PCGrad,wu2023_SFSC,Grad_Vaccine}. Formally, projecting $\bm g_i$ to the orthogonal plane $\bm g_j$ is expressed as:
\begin{equation}
\label{eq:proj}
\setlength{\abovedisplayskip}{4pt}
\setlength{\belowdisplayskip}{4pt}
\hat {\bm g}_i = \bm g_i - \frac{\bm g_i\cdot\bm g_j}{\parallel\bm g_j \parallel^2}\bm g_j,
\end{equation}
where $\cdot$ denotes the inner product. Considering that generally more than two gradient vectors are involved in optimization, we adopt an enumerate projection scheme to process all conflicting gradient vector pairs. We denote the gradient vectors after enumerate projection by $\{\hat {\bm g}_0, \hat {\bm g}_1,...,\hat {\bm g}_N\}$.

An intuitive observation is that the more a gradient vector conflicts with others, the larger the deviation between its projected direction and its original direction. Thus, we use the angle between a gradient vector and its projected counterpart to measure its conflicting degree with the others. Subsequently, we reweight the projected gradients based on the degree of conflict, thereby deriving an optimization direction in agreement with those of most loss functions. The conflict-aware gradient integration operation $\psi(\bm g_0, \bm g_1,..., \bm g_N)$ can be formulated as:
\begin{equation}
\label{eq:integration}
\setlength{\abovedisplayskip}{4pt}
\setlength{\belowdisplayskip}{4pt}
\begin{split}
&\tilde{\bm g} = \frac{\sum_{i=0}^{N}\left \langle\bm g_i,\hat {\bm g}_i\right \rangle^{\alpha}\bm g_i}{\sum_{i=0}^{N}\left \langle \bm g_i,\hat {\bm g}_i\right \rangle^{\alpha}}(N+1).
\end{split}
\end{equation}
Herein $\left \langle \cdot, \cdot \right \rangle$ calculates the cosine similarity between the inputs, and $\alpha$ is a hyperparameter controlling the influence of the conflicting degree on the weight. Algorithm 1 in \textbf{Appendix C} summarizes the optimization process.

Technically, we address gradient conflicts at a finer granularity, resolving them at the level of individual convolutional kernels and linear layers. Instead of flattening the gradients of all model parameters into a single vector, we process the flattened gradients of each convolutional kernel or linear layer individually with the above method.

\section{Experiment}
\begin{table*}[t]
\centering
\setlength{\tabcolsep}{10pt}
\renewcommand{\arraystretch}{0.95}
\caption{Comparisons on pre-determined capacities over GLDv2-test~\cite{gldv2}, RParis~\cite{rparis&roxford}, and ROxford~\cite{rparis&roxford}. We report the average of the mAP scores on the datasets. ResNet-18 is used as the backbone. $\phi_0$ denotes the dense network. The numerical subscript of a small-capacity (sub)network represents its capacity. We provide the detailed results for each dataset in \textbf{Appendix E}.}
\vspace{-2mm}
\resizebox{1.0\linewidth}{!}{
\begin{tabular}{l|ccccc|ccccc|ccccc}
\toprule
  \diagbox{$\phi_q$}{$\phi_g$}& $\phi_0$& $\phi_{80\%}$& $\phi_{60\%}$& $\phi_{40\%}$& $\phi_{20\%}$& $\phi_0$& $\phi_{80\%}$& $\phi_{60\%}$& $\phi_{40\%}$& $\phi_{20\%}$ & $\phi_0$& $\phi_{80\%}$& $\phi_{60\%}$& $\phi_{40\%}$& $\phi_{20\%}$\\
   \midrule
&\multicolumn{5}{c}{{Independent learning}} &\multicolumn{5}{c}{Joint learning} &\multicolumn{5}{c}{O2O-SSPL}\\
\midrule
    $\phi_0$& 45.41 & -- & -- & -- & -- & 43.94 & 43.55 & 43.44 & 42.36 & 42.23  & 45.41 & 43.70& 43.66 & 43.18 & 41.64 \\
    $\phi_{80\%}$& -- & 44.72& -- & -- & --  &43.27& 43.64&43.22 & 42.14 & 42.18 & 44.20 &  42.17&42.43 & 42.02& 40.50 \\
    $\phi_{60\%}$& -- & -- & 43.88 & -- & -- & 43.25 & 43.24 & 43.50& 42.44 & 42.43 &43.93 &41.85 &42.43 & 41.71&40.25 \\
    $\phi_{40\%}$& -- & -- & -- & 43.40 & -- &  42.69 & 42.79 & 42.84 & 42.24 & 41.51 & 43.63 & 42.09 & 42.00 & 41.77 & 40.14 \\
    $\phi_{20\%}$& -- & -- & -- & -- & 41.77 &42.58 & 42.51 & 42.69 & 41.19 & 41.86 & 42.59 &40.72&41.02 &40.42 & 39.69 \\
\midrule
&\multicolumn{5}{c}{{BCT-S w/ SwitchNet}} &\multicolumn{5}{c}{Asymmetric-S w/ SwitchNet}&\multicolumn{5}{c}{{SFSC}} \\

 \midrule
    $\phi_0$& 43.77 & 43.75& 43.37 & 42.80 &  41.77 & 45.09 & 33.11 &32.33 &  32.36 & 29.54& 44.47 & 44.39& 44.01& 43.54 & 42.45\\
    $\phi_{80\%}$&43.69  & 43.95 & 43.64 & 42.68 & 41.67 &33.72 & 30.39 &26.91& 26.75&24.16 & 44.40 & 44.28 & 43.90 & 43.55 & 42.47\\
    $\phi_{60\%}$&43.62  & 43.72 &43.53 & 42.48 & 41.39&  32.99&  27.74 & 28.75 & 26.80 & 24.40& 43.94 & 44.08 & 43.91 & 43.58 &  42.52 \\
    $\phi_{40\%}$&  43.08&42.99&  42.89 &42.68 & 42.38 &  31.96& 27.13 & 26.87 &28.50 & 24.95 &  43.67 & 43.57& 43.39 & 42.98& 41.92\\
    $\phi_{20\%}$& 42.33 &  42.24 &42.22 &41.55&40.70& 30.85& 25.16 & 25.42 &26.56& 25.93& 43.00 & 42.97 & 42.71 & 42.35 & 41.43 \\
\midrule
&\multicolumn{5}{c}{BCT-S w/ PrunNet} &\multicolumn{5}{c}{Asymmetric-S w/ PrunNet}&\multicolumn{5}{c}{{\textbf{Ours}}}\\
 \midrule
     $\phi_0$& 43.70 &43.70 & 43.72 & 43.58 &43.59&  45.17 & 45.21&45.09& 44.52 &42.94 & \textbf{46.29} & \textbf{46.29} &\textbf{46.30} & \textbf{46.27} & \textbf{46.08}\\
    $\phi_{80\%}$&43.68 & 43.72& 43.73 & 43.58& 43.58&45.30 &45.40 & 45.27& 44.57 & 42.94& \textbf{46.29} & \textbf{46.29} &\textbf{46.28}& \textbf{46.26} & \textbf{46.07}\\
    $\phi_{60\%}$&43.71 & 43.71 &43.71 & 43.58 &43.59 & 44.84 & 44.89&44.88 & 44.38& 42.64& \textbf{46.26} & \textbf{46.26}& \textbf{46.25} & \textbf{46.26} &\textbf{46.08} \\
    $\phi_{40\%}$&43.71 & 43.71& 43.71 & 43.59 & 43.60 & 44.13 & 44.15 & 44.25& 43.52&  42.27& \textbf{45.98} & \textbf{45.99} &\textbf{45.95}& \textbf{45.97} & \textbf{45.82}\\
    $\phi_{20\%}$& 43.59 & 43.59 & 43.58&  43.55& 43.57& 42.85 &43.10 & 43.05 &  42.64& 41.58& \textbf{45.61} & \textbf{45.63} &  \textbf{45.63} &\textbf{45.74} & \textbf{45.63} \\
\bottomrule
\end{tabular}}
\label{Tab:performance_landmark}
\vspace{-1mm}
\end{table*}

\begin{table*}[t]
\centering
\setlength{\tabcolsep}{3pt}
\renewcommand{\arraystretch}{1.0}
\caption{Comparisons on pre-determined capacities over In-shop~\cite{inshop}.  We use ResNet-18 as the backbone and report the Recall@1 score.}
\vspace{-2mm}
\resizebox{1.0\linewidth}{!}{
\begin{tabular}{l|c|cc|cc|cc|cc}
\toprule
 & $\mathcal{M}(\phi_0,\phi_0)$ & $\mathcal{M}(\phi_{80\%},\phi_{80\%})$ & $\mathcal{M}(\phi_{80\%},\phi_0)$ & $\mathcal{M}(\phi_{60\%},\phi_{60\%})$ & $\mathcal{M}(\phi_{60\%},\phi_0)$ & $\mathcal{M}(\phi_{40\%},\phi_{40\%})$ & $\mathcal{M}(\phi_{40\%},\phi_0)$ & $\mathcal{M}(\phi_{20\%},\phi_{20\%})$ & $\mathcal{M}(\phi_{20\%},\phi_0)$\\
 & Self-test & Self-test & Cross-test & Self-test & Cross-test & Self-test & Cross-test & Self-test & Cross-test\\
\midrule
    Independent learning & 86.14 & 85.51 &-- &84.75&--&84.48&--&83.51&-- \\
    SFSC&84.57  & 84.48 & 84.40&84.25&84.31&84.15&84.20&83.57&83.74 \\
    \textbf{Ours} & \textbf{87.31} & \textbf{87.30} & \textbf{87.33} & \textbf{87.21}&\textbf{87.23}&\textbf{87.14}&\textbf{87.15}&\textbf{86.43}&\textbf{86.77} \\
\bottomrule
\end{tabular}}
\label{Tab: inshop results}
\vspace{-2mm}
\end{table*}

\subsection{Experimental settings}

\noindent\textbf{Benchmarks.}
We evaluate PrunNet on the landmark benchmarks (GLDv2~\cite{gldv2}, RParis~\cite{rparis&roxford}, and ROxford~\cite{rparis&roxford}), the commodity benchmark (In-shop~\cite{inshop}), and the ReID benchmark (VeRi-776~\cite{VeRi776}).
GLDv2 contains 1,580,470 images from 81,313 landmarks. We use a subset of GLDv2 containing 24,393 classes to train the model to reduce training resource consumption, and evaluate the model on GLDv2-test, RParis, and ROxford. In-shop consists of 52,712 images of 7,982 clothing items. VeRi-776 is a vehicle ReID dataset containing 51,035 images from 776 vehicles.

\vspace{1mm}
\noindent\textbf{Metrics.}
We denote the evaluation metric of retrieval performance as $\mathcal{M}(\phi_q, \phi_g)$, where $\phi_q$ and $\phi_g$ are the query and gallery models used to extract query and gallery embeddings, respectively. In \textbf{self-test} where $\phi_q$ and $\phi_g$ are the same model, $\mathcal{M}(\phi_q, \phi_g)$ measures the discriminability of this model. In contrast, in \textbf{cross-test}, $\mathcal{M}(\phi_q, \phi_g)$ assesses the compatibility between $\phi_q$ and $\phi_g$. Besides, mAP is used as the metric for the landmark and ReID benchmarks, while Recall@1 is used as the metric for In-shop.

\vspace{1mm}
\noindent\textbf{Implementation details.}
We use various network architectures to implement PrunNet, including ResNet~\cite{Resnet}, MobileNet V2~\cite{Mobilenetv2}, ResNeXt~\cite{Resnext} and ViT~\cite{ViT}. A linear layer is appended to the backbone to convert the feature dimension to 256. When performing backward propagation of each subnetwork, we filter the gradient of $s^{l}_{ij}$ with $r(s^{l}_{ij})$ to eliminate the influence on the pruned connections. Practically, the mean and variance of the Batch Normalization (BN) layers differ substantially across subnetworks of varying capacities. We employ Adaptive BN~\cite{Adaptive_BN} to recalculate the mean and variance for each subnetwork after model training. By default, $N$ is set to 4, and the capacities of subnetworks are set to 20\%, 40\%, 60\%, and 80\% unless otherwise specified. Following~\cite{wu2023_SFSC, BCT}, we impose the compatible constraint in the logit space. Specifically, we append a classifier on the PrunNet and employ the cross-entropy loss to serve as $\{\mathcal{L}_0,\mathcal{L}_1,...,\mathcal{L}_N\}$. Please refer to \textbf{Appendix B} for more implementation details. 

\subsection{Comparisons on pre-determined capacities}
We begin with the experiment that simulates building retrieval models for pre-determined platforms. We assess the models at pre-defined capacities obtained by these methods:

\noindent\textbf{Independent learning}, where networks at different capacities are trained independently using the cross-entropy loss;

\noindent\textbf{Joint leaning}, where independent networks sharing a common classifier are trained with the combined cross-entropy loss applied to each model.

\noindent\textbf{One-to-one compatible learning (O2O-SSPL)}, where the small networks are trained to align with the dense network by the recently proposed SSPL~\cite{wu2023_SSPL}.

\noindent\textbf{SFSC}~\cite{wu2023_SFSC}, which trains a SwitchNet containing pre-defined subnetworks. We reproduce this method following the paper, as its source codes have not been released.

\noindent\textbf{BCT-S/Asymmetric-S with SwitchNet}~\cite{wu2023_SFSC}, directly training SwitchNet using the combined cross-entropy or contrastive~\cite{budnik2021_AML} loss applied to each subnetwork, respectively.

\noindent\textbf{BCT-S/Asymmetric-S with PrunNet}, training PrunNet like the above two methods.

\subsubsection{Results on various benchmarks}
Table~\ref{Tab:performance_landmark} presents the average mAP on GLDv2-test, RParis, and ROxford. Table~\ref{Tab: inshop results} and Table~\ref{Tab: veri776 results} report the results on In-shop and VeRi-776, respectively. Specifically, we use the same setting of the subnetwork capacities as SFSC~\cite{wu2023_SFSC} on VeRi-776, so that we can directly include the results reported in~\cite{wu2023_SFSC} in the comparison. We can observe that our reproduced results closely align with the official results. 

\begin{table*}[h!]
\centering
\setlength{\tabcolsep}{6pt}
\renewcommand{\arraystretch}{0.95}
\caption{Comparisons on pre-determined capacities over Veri-776~\cite{VeRi776}. We employ ResNet-18 as the backbone. We use the same setting for the subnetwork capacities as SFSC~\cite{wu2023_SFSC} to include the results reported by~\cite{wu2023_SFSC} (denoted by $\dagger$) in the comparison on Veri-776.}
\vspace{-2mm}
\resizebox{1.0\linewidth}{!}{
\begin{tabular}{l|c|cc|cc|cc}
\toprule
 & $\mathcal{M}(\phi_0,\phi_0)$ & $\mathcal{M}(\phi_{56.25\%},\phi_{56.25\%})$ & $\mathcal{M}(\phi_{56.25\%},\phi_0)$ & $\mathcal{M}(\phi_{25\%},\phi_{25\%})$ & $\mathcal{M}(\phi_{25\%},\phi_0)$ & $\mathcal{M}(\phi_{6.25\%},\phi_{6.25\%})$ & $\mathcal{M}(\phi_{6.25\%},\phi_0)$ \\
 & Self-test & Self-test & Cross-test & Self-test & Cross-test & Self-test & Cross-test \\
\midrule
    Independent learning &66.57&56.91&--&53.15&--&44.40&-- \\
    SFSC$^{\dagger}$ & 66.55&--&62.72&--&62.28&--&55.04\\
    SFSC &66.11&62.62&65.35&58.13&63.22&50.34&57.94 \\
    \textbf{Ours} &\textbf{67.82}& \textbf{67.11} &\textbf{67.58}&\textbf{64.45} &\textbf{66.25}&\textbf{54.45}&\textbf{58.30}\\
\bottomrule
\end{tabular}}
\label{Tab: veri776 results}
\vspace{-1mm}
\end{table*}

\begin{table*}[h!]
\centering
\setlength{\tabcolsep}{2pt}
\renewcommand{\arraystretch}{0.95}
\caption{Comparisons on pre-determined capacities over GLDv2-test~\cite{gldv2}, RParis~\cite{rparis&roxford}, and ROxford~\cite{rparis&roxford} using different backbones. We report the average mAP score on the three datasets.}
\vspace{-2mm}
\resizebox{1.0\linewidth}{!}{
\begin{tabular}{ll|c|cc|cc|cc|cc}
\toprule
&& $\mathcal{M}(\phi_0, \phi_0)$ & $\mathcal{M}(\phi_{80\%}, \phi_{80\%})$ & $\mathcal{M}(\phi_{80\%}, \phi_0)$ & $\mathcal{M}(\phi_{60\%}, \phi_{60\%})$ & $\mathcal{M}(\phi_{60\%}, \phi_0)$ & $\mathcal{M}(\phi_{40\%}, \phi_{40\%})$ & $\mathcal{M}(\phi_{40\%}, \phi_0)$ & $\mathcal{M}(\phi_{20\%}, \phi_{20\%})$ & $\mathcal{M}(\phi_{20\%}, \phi_0)$\\
&& Self-test & Self-test & Cross-test & Self-test & Cross-test & Self-test & Cross-test & Self-test & Cross-test\\
\midrule
    \multirow{3}{*}{ResNet-50}& Independent learning & 47.06  & 46.84 & -- & 46.47 & -- & 46.34 & -- & 44.74 & --\\
    &SFSC &46.42&46.37&46.42&46.19&46.36&46.11&46.17&45.48& 45.84\\
    &\textbf{Ours}& \textbf{47.88} &\textbf{47.74}& \textbf{47.79} &\textbf{47.72} & \textbf{47.79}& \textbf{47.58} & \textbf{47.73}&\textbf{47.22}&\textbf{47.61} \\
\midrule
    \multirow{3}{*}{ResNeXt-50}& Independent learning & 47.84&46.90 & -- & 46.26 & -- & 45.78 & -- & 43.88&-- \\
    &SFSC &47.09&46.36&46.24&45.85&45.98&45.11&45.74&43.57&45.16\\
    &\textbf{Ours} & \textbf{48.90} &\textbf{48.92}&\textbf{48.91}&\textbf{48.96}&\textbf{48.89}&\textbf{49.01}& \textbf{48.92} &  \textbf{48.21}&\textbf{48.58} \\
\midrule
    \multirow{3}{*}{MobileNet-V2}&Independent learning& 40.53& 39.87& -- & 39.44 & -- & 38.52 & -- &37.95&-- \\
    &SFSC&40.24 &39.66&39.58&39.83&39.97&39.19&39.64&37.76&38.49\\
    &\textbf{Ours} &\textbf{41.19}  &\textbf{41.27} & \textbf{41.18} & \textbf{41.29}&\textbf{41.22} &\textbf{40.72}& \textbf{41.03}& \textbf{38.97} &\textbf{40.16}\\
\midrule
    \multirow{3}{*}{ViT-Small}& Independent learning & 51.91 & 46.71 &--   & 42.06 & -- & 35.75 & -- & 28.16&-- \\
    &SFSC& 48.96&45.89&47.39&42.89&46.13&39.72&43.21&29.84&35.34\\
    &\textbf{Ours} & \textbf{52.39} & \textbf{50.10} & \textbf{50.31}& \textbf{49.80}& \textbf{50.34}&\textbf{47.39} &\textbf{48.05} &\textbf{41.21} & \textbf{43.40} \\
\bottomrule
\end{tabular}}
\label{Tab: different model architecture}
\vspace{-2mm}
\end{table*}

On these benchmarks, our algorithm achieves the best self-test and cross-test performance. The learned multi-prize subnetworks as well as the dense network by our methods outperform the models trained independently at the same capacities. Additionally, it is intriguing that some prize subnetworks outperform the dense network slightly, which also has been observed in the LTH studies~\cite{What's_hidden,MPT}. This can be attributed to the reduction of unnecessary redundant weights, enabling the network to focus more effectively on essential information for tasks. We present more results on additional benchmarks in \textbf{Appendix F}.

\subsubsection{Results using various backbones}
We also assess our method on several representative visual backbones, including ResNet-50~\cite{Resnet}, ResNeXt-50~\cite{Resnext}, MobileNet-V2~\cite{Mobilenetv2}, and ViT-Small~\cite{ViT}. Particularly, we use the proposed prunable linear layer to implement the attention block and feedforward block of ViT-Small. 
Table~\ref{Tab: different model architecture} compares the performance of our method and SFSC~\cite{wu2023_SFSC} on GLDv2-test, RParis, and ROxford. The superior performance of our method across various network architectures demonstrates its generalizability.

\subsection{Comparisons on new capacities}

\begin{table*}[h!]
\centering
\setlength{\tabcolsep}{14pt}
\renewcommand{\arraystretch}{0.95}
\caption{Comparisons on a new capacity (10\%) over GLDv2-test~\cite{gldv2}, RParis~\cite{rparis&roxford}, and ROxford~\cite{rparis&roxford}. We report the average mAP score on the three datasets. ResNet-18 is used as the backbone. For methods without PrunNet, we use BCT~\cite{BCT} or SSPL~\cite{wu2023_SSPL} to train a new small-capacity model, which retains 10\% of the weights from the dense network $\phi_0$, with compatibility with existing models.}
\vspace{-2mm}
\resizebox{1.0\linewidth}{!}{
\begin{tabular}{l|ccccc|c}
\toprule
& $\mathcal{M}(\phi_{10\%}, \phi_0)$ & $\mathcal{M}(\phi_{10\%}, \phi_{80\%})$ & $\mathcal{M}(\phi_{10\%}, \phi_{60\%})$& $\mathcal{M}(\phi_{10\%}, \phi_{40\%})$ & $\mathcal{M}(\phi_{10\%}, \phi_{20\%})$ & $\mathcal{M}(\phi_{10\%}$, $\phi_{10\%})$ \\
\midrule
Joint learning + BCT & 43.13 & 43.06 & 42.25 & 42.09 & 41.78 & 41.28\\
O2O-SSPL + SSPL & 41.49 &40.18&40.24&39.81&38.24&38.10 \\
\midrule
BCT-S w/ SwitchNet + BCT &41.71&41.64&41.46&40.79&39.78&40.01\\
Asymmetric-S w/ SwitchNet + BCT &42.57&32.03&31.25&31.19&28.62&40.02 \\
SFSC + BCT &41.59&41.54&41.37&40.92&39.79&39.56\\
\midrule
BCT-S w/ PrunNet&42.10&42.09&42.09&42.08&42.04&40.32\\
Asymmetric-S w/ PrunNet&37.58&37.73&37.73&37.69&37.04 &34.22\\
\textbf{Ours} &\textbf{44.67}&\textbf{44.63}&\textbf{44.66}&\textbf{44.72}&\textbf{44.55}&\textbf{42.55}\\
\bottomrule
\end{tabular}}
\label{Tab:new capacities}
\end{table*}

\begin{figure}[t!]
\centering
\includegraphics[width=0.99\linewidth]{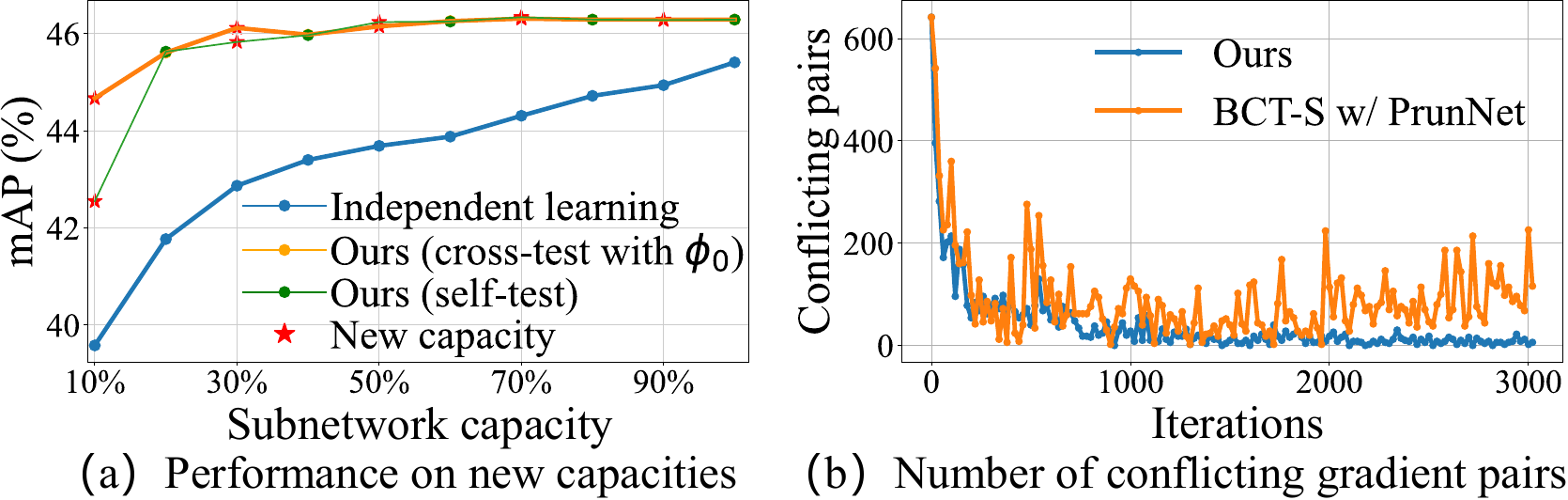}
\vspace{-2mm}
\caption{(a) The performance of our method at new capacities. (b) The number of conflicting gradient pairs in the first convolutional layer of PrunNet. ResNet-18 is used as the backbone.}
\label{Fig:new capacity and conflicts}
\vspace{-2mm}
\end{figure}

We also conduct experiments simulating the deployment demand on new platforms, requiring compatible models at novel capacities. For the methods using independent networks or SwitchNet, we leverage BCT~\cite{BCT} to learn a compatible model at the desired capacity. Particularly, for O2O-SSPL~\cite{wu2023_SSPL}, we still use SSPL to train a model at the desired capacity. Assuming the desired capacity is 10\% of the dense network, Table~\ref{Tab:new capacities} shows the experimental results on landmark benchmarks. Our method achieves the best performance in both self-test of $\phi_{10\%}$ and the cross-test with existing subnetworks. Additionally, we assess our method at more novel capacities, as shown in Figure~\ref{Fig:new capacity and conflicts} (a). Our method outperforms independent learning models while maintaining high compatibility with the dense network, demonstrating its effectiveness in satisfying new deployment demands.

\subsection{Ablation studies}
We conduct experiments to investigate the effect of the core designs of our PrunNet on the landmark benchmarks. 

\begin{table*}[h!]
\centering
\setlength{\tabcolsep}{8pt}
\renewcommand{\arraystretch}{0.95}
\caption{Results of different variants on GLDv2-test~\cite{gldv2}, RParis~\cite{rparis&roxford}, and ROxford~\cite{rparis&roxford}. We report the average mAP score of the datasets.}
\label{Tab:ablation}
\vspace{-2mm}
\resizebox{1.0\linewidth}{!}{
\begin{tabular}{l|c|cc|cc|cc|cc}
\toprule
& $\mathcal{M}(\phi_0, \phi_0)$ & $\mathcal{M}(\phi_{80\%}, \phi_{80\%})$ & $\mathcal{M}(\phi_{80\%}, \phi_0)$ & $\mathcal{M}(\phi_{60\%}, \phi_{60\%})$ & $\mathcal{M}(\phi_{60\%}, \phi_0)$ & $\mathcal{M}(\phi_{40\%}, \phi_{40\%})$ & $\mathcal{M}(\phi_{40\%}, \phi_0)$ & $\mathcal{M}(\phi_{20\%}, \phi_{20\%})$ & $\mathcal{M}(\phi_{20\%}, \phi_0)$\\
& Self-test & Self-test & Cross-test & Self-test & Cross-test & Self-test & Cross-test & Self-test & Cross-test\\
\midrule
    Independent learning & 45.41 & 44.72 & --&43.88&--&43.40&--&41.77&-- \\
\midrule
    \textbf{Ours} ($N$ = 4) & 46.29 & 46.29& 46.29 & 46.25&46.26&45.97&45.98&\textbf{45.63}&45.61 \\
    Frozen scores &45.23&45.11&45.18&45.01&45.09&44.69&45.00&43.00&43.96\\
    $N$ score maps ($N$ = 4) &44.26&43.96&43.91&43.64&43.62&43.22&43.72&42.23&43.05\\
    \midrule
    Direct gradient integration& 45.70&45.67&45.68&45.59&45.68&45.79&45.65&45.30&45.44\\
    Direct loss combination &43.55&43.43&43.51&43.10&43.27&42.80&43.08&42.14&42.53\\
    Pareto integration &44.84&44.80&44.84&44.80&44.83&44.70&44.82& 43.84&44.49\\
    Without weight optimization&3.08&3.19&2.49&3.21&2.41&3.14&2.38&2.99&2.24\\
    \midrule
   \textbf{Ours} ($N$ = 1)&44.85  & 44.93 &44.80&44.82&44.74& 44.73&44.79&40.85&42.21 \\
    \textbf{Ours} ($N$ = 2)&45.63  & 45.56 & 45.56&45.55&45.57&45.39&45.53&42.40&44.59 \\
    \textbf{Ours} ($N$ = 6)& \textbf{46.33} &\textbf{46.31} & \textbf{46.31}&\textbf{46.27}& \textbf{46.30}&\textbf{46.03}&\textbf{46.18}&45.40&\textbf{45.92}\\
\bottomrule
\end{tabular}}
\vspace{-1mm}
\end{table*}

\vspace{1mm}
\noindent\textbf{Effect of the learnable scores.}
We analyze the effect of the learnable scores by training a variant whose scores are frozen. It means that the architecture of subnetworks is pre-defined by the initial score values. As shown in Table~\ref{Tab:ablation}, freezing the score leads to large performance drops for the dense network and all subnetworks, which demonstrates that optimizing both the architecture and weight of subnetworks benefits finding multi-prize subnetworks.

\vspace{1mm}
\noindent\textbf{Effect of greedy pruning.}
We further construct a prunable network with $N$ learnable score maps, each corresponding to a pre-defined capacity. The weights of each subnetwork can be selected from the entire dense network, expanding the search space for subnetwork architectures. Nevertheless, it complicates the model optimization and affects the performance adversely, as shown in Table~\ref{Tab:ablation}. The result validates the effectiveness of the greedy pruning mechanism.

\vspace{1mm}
\noindent\textbf{Effect of the proposed optimization method.}
We evaluate four variants to analyze the effect of our optimization method: 1) Direct loss combination, minimizing the summation of the cross-entropy losses; 2) Pareto integration, using the popular Pareto algorithm~\cite{Multi_pareto} to process the gradient of each loss; 3) Direct gradient integration, which replaces the conflict-aware gradient integration,~\ie, Eq.~\eqref{eq:integration}, with direct summation integration while retaining the enumerate projection; and 4) Keeping the weights frozen and optimizing the scores alone. As shown in Table~\ref{Tab:ablation}, our approach outperforms these three variants, demonstrating the effectiveness of our conflict-aware gradient integration.

\vspace{1mm}
\noindent\textbf{Analyses on hyperparameter $N$.}
We also assess our method with varying $N$ during training, as shown in Table~\ref{Tab:ablation}. Using one subnetwork ($N=1$) hinders learning accurate weight ranking, leading to a large performance drop for the sparse subnetwork $\phi_{20\%}$. By contrast, using more subnetworks benefits learning more accurate weight ranking and contributes to better performance. 

\subsection{Visualizations and analyses}

\noindent\textbf{Number of conflicting gradient pairs.}
Figure~\ref{Fig:new capacity and conflicts} (b) shows the number of conflicting gradient pairs encountered during PrunNet (ResNet-18) optimization using our method and BCT-S. In this analysis, we count the convolutional kernels with conflicting gradient vectors across different losses in the first convolutional layer of the backbone. We observe that both our method and BCT-S encounter numerous conflicts at the beginning. However, when using our proposed learning approach, the number of conflicts significantly decreases and remains at a low level, which indicates that our method fosters more stable network convergence. 

\begin{figure}[t!]
\centering
\includegraphics[width=0.98\linewidth]{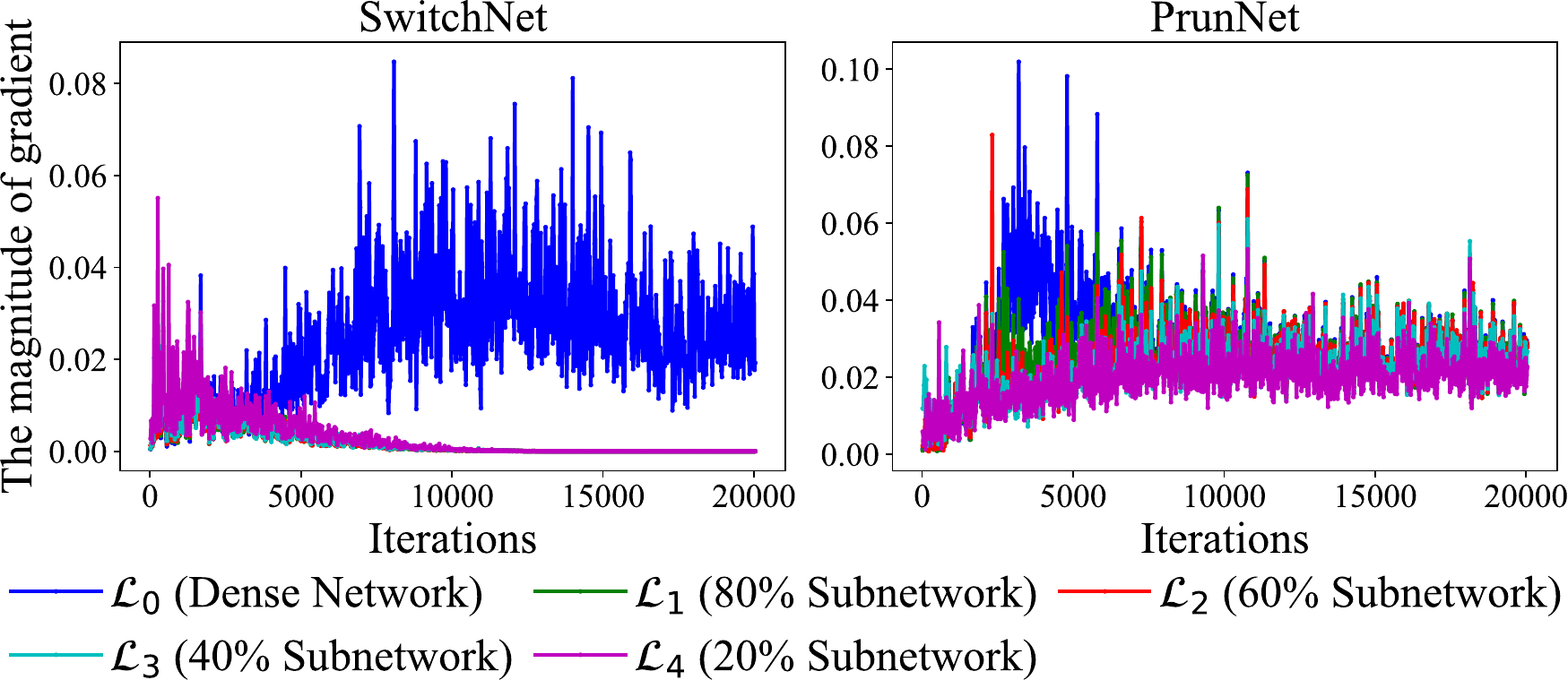}
\vspace{-1mm}
\caption{The gradient magnitudes of a convolutional kernel in SwitchNet and PrunNet when optimizing them with our losses. The gradient magnitudes of PrunNet exhibit consistency across different losses along with the training progress.}
\label{Fig:Visualize grad magnitude}
\vspace{-2mm}
\end{figure}

\vspace{1mm}
\noindent\textbf{Analyses on the gradient amplitude.}
Several MTL studies~\cite{wu2023_SFSC,PCGrad,Gradnorm} have observed the gradient magnitude discrepancies that affect model optimization. We examine the gradient magnitudes of a convolutional kernel in PrunNet and SwitchNet when optimizing them with our losses. As shown in Figure~\ref{Fig:Visualize grad magnitude}, the gradient magnitudes of PrunNet exhibit consistency across different losses, while those of SwitchNet do not. We attribute this phenomenon to that the magnitude of a gradient vector in PrunNet is primarily influenced by high-scoring weights. This comparison suggests that PrunNet is easier to train and more stably convergent.

\section{Conclusion}
In this paper, we propose a prunable network that can generate compatible multi-prize subnetworks at different capacities for multi-platform deployment. Specifically, we optimize the weight and architectures of the multi-prize subnetworks within a dense network simultaneously using our proposed conflict-aware gradient integration scheme. Our method achieves state-of-the-art performances on diverse retrieval benchmarks. We will explore implementing our idea through structured pruning in future work, which is more friendly for acceleration than unstructured pruning.

\section{Acknowledgement}
This work was supported in part by the National Natural Science Foundation of China (Grant No. 62372133, 62125201, U24B20174), in part by Shenzhen Fundamental Research Program (Grant No. JCYJ20220818102415032).
\clearpage
\setcounter{page}{1}
\appendix
\maketitlesupplementary

\counterwithin{table}{section}
\counterwithin{figure}{section}

\renewcommand{\thetable}{\Alph{table}}
\renewcommand{\thefigure}{\Alph{figure}}

\section{Additional details of weight inheritance}
\vspace{2mm}

We briefly present our preliminary experiment in the main manuscript. Herein we provide more details and analyses. We perform pruning with the edge-popup algorithm on an 8-layer convolutional network following ~\cite{What's_hidden}. Specifically, we attach a learnable score to each randomly initialized weight of the network, keeping the weight frozen while updating the score to discover a good subnetwork during training. We explore two pruning strategies, One-shot Pruning (OSP) and Iterative Pruning (IP) in our preliminary experiment. OSP proposed in~\cite{What's_hidden} is employed as the control group, and IP is introduced to investigate the weight inheritance nature of multi-prize subnetworks. 

As presented in~\cite{What's_hidden}, the subnetwork discovered by OSP at a capacity of 50\% achieves the best performance among all the subnetworks with various capacities. Thus, we begin with a subnetwork at the capacity of 50\% to perform iterative pruning. For example, we identify a well-performing 40\%-subnetwork from the 50\%-subnetwork and repeat this process in a greedy pruning manner to progressively obtain subnetworks of varying capacities. As illustrated in Figure 2 in the main manuscript, the subnetworks identified by IP outperform those obtained by OSP. It empirically demonstrates that small-capacity prize subnetwork can be obtained by selectively inheriting weights from a large-capacity prize subnetwork, rather than searching for it within the entire dense network. 

For the rationale behind the weight inheritance nature, we speculate that connections within a network exhibit varying degrees of importance. Integrating a set of critical connections is essential for identifying a well-performing subnetwork. The performance of a highly sparse subnetwork can be enhanced by adding an appropriate number of connections until redundancy arises. Furthermore, when attempting to directly identify a highly sparse subnetwork using the OSP method, critical connections are often excluded prematurely during the early training stages due to incomplete convergence of the learned scores. This explains why OSP tends to be less effective than the IP approach for identifying sparse subnetworks.

\begin{algorithm}[t!]
\caption{Training process of our method}
\label{pseudo algorithm}
\SetKwInOut{Require}{Require}
\SetAlgoLined
\LinesNumbered
\SetNoFillComment 
\SetCommentSty{} 
\SetSideCommentRight 
\Require{Batch input $\mathcal{B}$, the dense model $\phi_0$, model parameters $\theta$, $N$ capacity factors $\{c_i\}_{i=1}^{N}$; }
\tcp{\small \color{gray} Backward propagation}
$\bm{g}_0 \gets \frac{\partial \mathcal{L}_0(\phi_0,\mathcal{B})}{\partial \phi_0}$\;
\For{$c_i \in \{c_i\}_{i=1}^{N}$}{
    $\phi_i \gets$ GetSubmodel($\phi_0,c_i$)\;
    $\bm{g}_i \gets \frac{\partial \mathcal{L}_i(\phi_i,\mathcal{B})}{\partial \phi_i}$\;
}
 $G,G_{ori} \gets \{\bm{g}_0,\bm{g}_1,...,\bm{g}_N\}$\;
\tcp{\small \color{gray} Conflict-aware gradient integration}
\For{$\bm{g}_i \in G$}{
    $G' \gets$ Shuffle$(G)$\;
    \For{$\bm{g}_j \in G'$}{
        \If{$ \bm{g}_i \cdot \bm{g}_j < 0$}{
            $\bm{g}_i \gets \bm{g}_i - \frac{\bm{g}_i\cdot \bm{g}_j}{\|\bm{g}_j\|^2}\bm{g}_j$ \;
        }
    }
}
\tcp{\small \color{gray} Calculate the cosine similarities}
\For{$\hat {\bm g}_i, \bm{g}_k \in G, G_{ori}$}{
    $\gamma_i \gets \left \langle\bm{g}_k,\hat {\bm g}_i\right \rangle^{\alpha}$\; 
}
$\bm{\tilde{g}} \gets \sum \frac{\gamma_i \hat {\bm g}_i}{\sum \gamma_i}(N+1)$ \;
\Return Update $\phi_0$ by $\Delta \theta$ = $\bm{\tilde{g}}$
\end{algorithm}

\begin{figure*}[t!]
\centering
\includegraphics[width=0.925\linewidth]{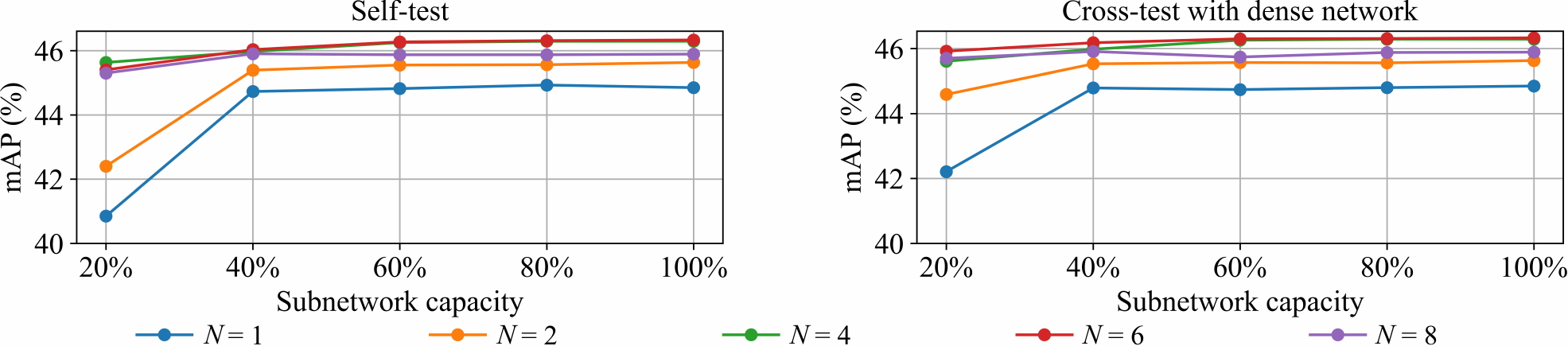}
\vspace{-1mm}
\caption{Performance of our PrunNet when different numbers of pre-defined subnetworks are used for modeling training. We show the average mAP of RParis~\cite{rparis&roxford}, ROxford~\cite{rparis&roxford}, and GLDv2-test~\cite{gldv2}. The cross-test values at 100\% capacity are identical to those of the self-test.}
\label{Fig:Model Number}
\vspace{-1mm}
\end{figure*}

\begin{figure*}[t!]
\centering
\includegraphics[width=0.925\linewidth]{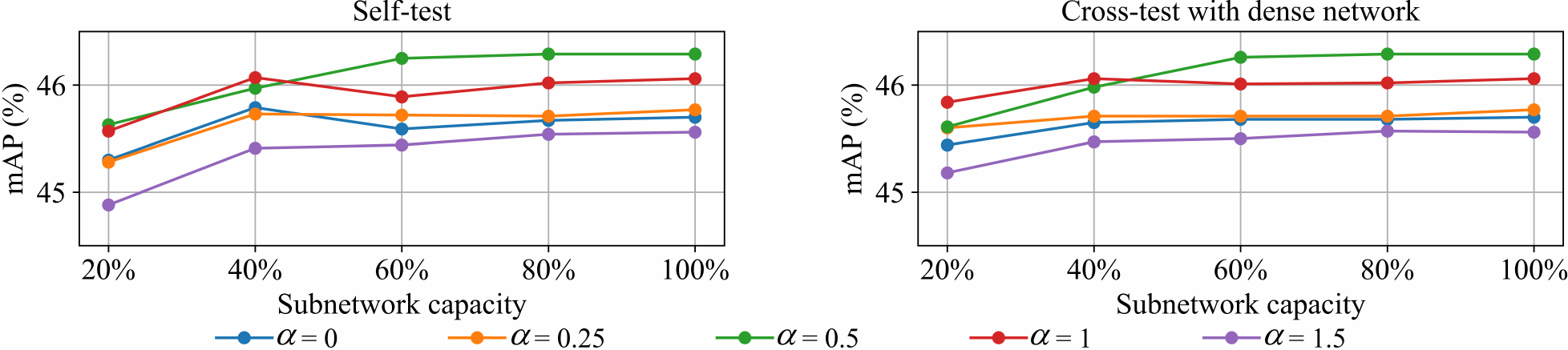}
\vspace{-1mm}
\caption{Performance across different values of $\alpha$ in Eq. (5) in the main manuscript. We show the average mAP of RParis~\cite{rparis&roxford}, ROxford~\cite{rparis&roxford}, and GLDv2-test~\cite{gldv2}. The cross-test values at 100\% capacity are identical to those of the self-test. When $\alpha$ is set to 0, our method is simplified to direct gradient integration after projection.} 
\label{Fig:hyperparameter_alpha}
\vspace{-1mm}
\end{figure*}

\begin{figure*}[h!]
\centering
\includegraphics[width=0.925\linewidth]{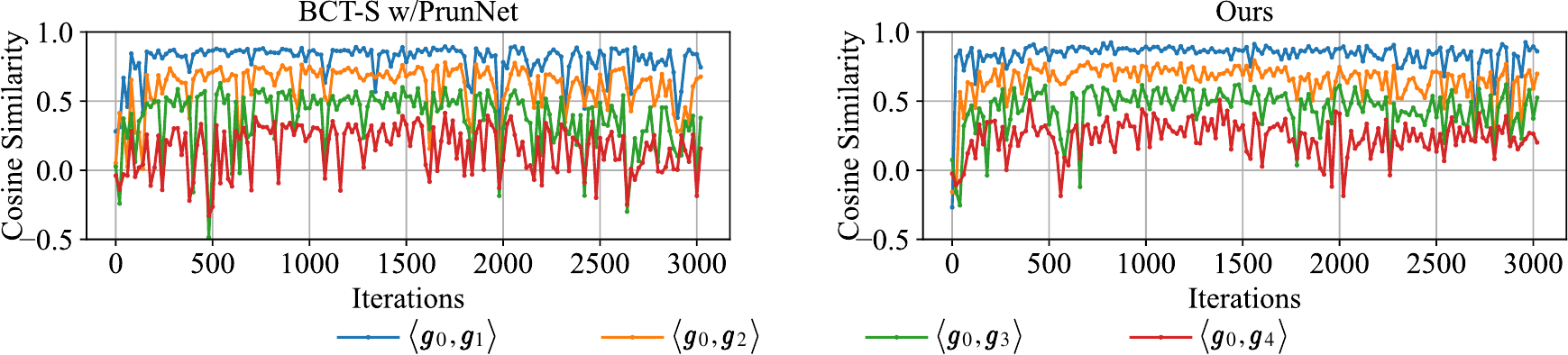}
\vspace{-1mm}
\caption{Cosine similarities between the gradient vectors of a single convolutional kernel in the dense network and each subnetwork when training PrunNet on GLDv2~\cite{gldv2}. ResNet-18 is used as the backbone. Herein $\bm g_0$ denotes the gradient vector of a convolutional kernel of the dense network, while $\bm g_1$, $\bm g_2$, $\bm g_3$, and $\bm g_4$ represent those of the subnetworks $\phi_{80\%}$, $\phi_{60\%}$, $\phi_{40\%}$, and $\phi_{20\%}$, respectively. $\left \langle \cdot, \cdot \right \rangle$ denotes the cosine similarity. 
The gradient vector of each subnetwork conflicts with that of the dense network at the beginning of the training. As training progresses, negative cosine similarities in our method occur only occasionally. In contrast, the subnetworks trained with the BCT-S method encounter negative cosine similarities more frequently.}
\label{Fig:Visualize cos similarity}
\vspace{-1mm}
\end{figure*}

\begin{figure*}[t!]
\centering
\includegraphics[width=0.8\linewidth]{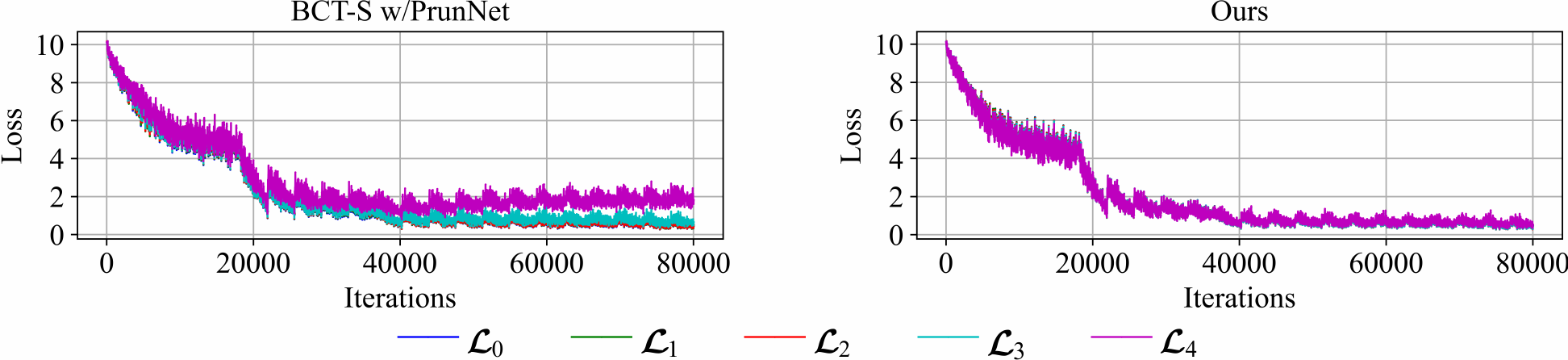}
\vspace{-1mm}
\caption{Loss convergence curves when training PrunNet with our method and BCT-S on GLDv2~\cite{gldv2}. $\mathcal{L}_0$ denotes the loss of dense network $\phi_0$, $\mathcal{L}_1$, $\mathcal{L}_2$, $\mathcal{L}_3$, and $\mathcal{L}_4$ denote the loss of the subnetworks $\phi_{80\%}$, $\phi_{60\%}$, $\phi_{40\%}$, and $\phi_{20\%}$, respectively. ResNet-18 is used as the backbone. The loss for both methods declines sharply at the beginning. However, as training progresses, BCT-S struggles to further reduce the losses of subnetworks. In contrast, the losses of all networks remain consistent and converge to lower values when using our method.}
\label{Fig:Visualize loss scale}
\vspace{-1mm}
\end{figure*}

\begin{figure}[t!]
\centering
\includegraphics[width=0.925\linewidth]{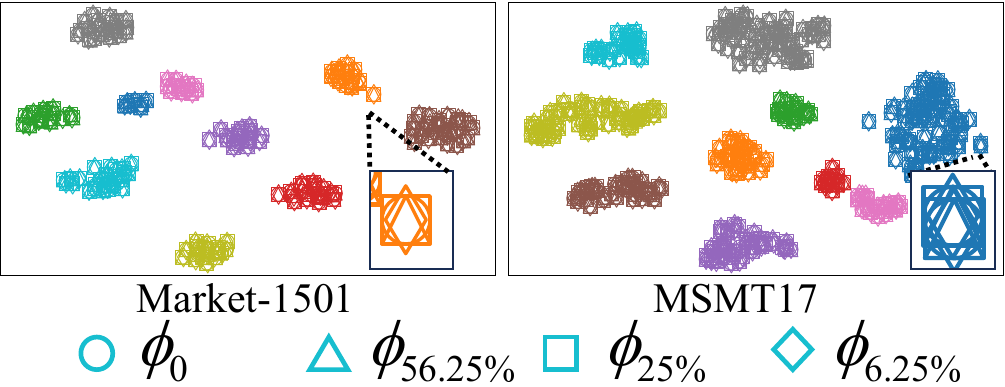}
\caption{Feature distributions of different capacities of subnetworks on Market-1501 and MSMT17 datasets visualized with t-SNE. Herein we randomly sample ten different persons on each dataset. We can observe that the feature distributions of subnetworks are aligned with that of the dense network, validating the compatibility among subnetworks.}
\label{Fig:feature_distribution}
\vspace{-1mm}
\end{figure}

\begin{table*}[t!]
\centering
\setlength{\tabcolsep}{9pt}
\renewcommand{\arraystretch}{0.95}
\caption{Detailed comparisons on pre-determined capacities over RParis~\cite{rparis&roxford}, ROxford~\cite{rparis&roxford} and GLDv2-test~\cite{gldv2}. ResNet-18 is used as the backbone. $\phi_0$ denotes the dense network. The numerical subscript of a small-capacity (sub)network represents its capacity.}
\resizebox{1.0\linewidth}{!}{
\begin{tabular}{l|ccccc|ccccc|ccccc}
\toprule
  \diagbox{$\phi_q$}{$\phi_g$}& $\phi_0$& $\phi_{80\%}$& $\phi_{60\%}$& $\phi_{40\%}$& $\phi_{20\%}$& $\phi_0$& $\phi_{80\%}$& $\phi_{60\%}$& $\phi_{40\%}$& $\phi_{20\%}$ & $\phi_0$& $\phi_{80\%}$& $\phi_{60\%}$& $\phi_{40\%}$& $\phi_{20\%}$\\
   \midrule
\rowcolor[gray]{0.9}\multicolumn{16}{c}{\emph{RParis}}\\
\midrule

\multicolumn{1}{c}{}&\multicolumn{5}{c}{{Independent learning}} &\multicolumn{5}{c}{Joint learning} &\multicolumn{5}{c}{O2O-SSPL}\\
\midrule
    $\phi_0$& 73.35 & -- & -- & -- & -- & 71.58 & 70.94 & 70.72 & 69.88 & 69.11  & 73.35 & 71.94 & 71.50 & 71.08 & 69.40 \\
    $\phi_{80\%}$& -- & 71.84& -- & -- & --  &71.53 & 71.50& 71.08 & 70.14 & 69.49 & 72.16 & 70.71 &70.19 & 69.85& 68.25 \\
    $\phi_{60\%}$& -- & -- & 70.71 & -- & -- & 71.22 & 70.66 & 70.75& 70.58 & 69.69 &71.91  &70.58 & 70.26 & 69.74 &68.26 \\
    $\phi_{40\%}$& -- & -- & -- & 70.37 & -- & 70.65 & 70.28 & 70.10 & 69.28 & 68.56 & 71.96 & 70.59 & 70.18 & 69.89 & 68.14 \\
    $\phi_{20\%}$& -- & -- & -- & -- & 67.77 &70.79 & 70.24 & 70.05 & 68.80 & 68.80 & 70.19 &68.94&68.32 &68.14 & 66.72 \\
\midrule
\multicolumn{1}{c}{}&\multicolumn{5}{c}{{BCT-S w/ SwitchNet}} &\multicolumn{5}{c}{Asymmetric-S w/ SwitchNet}&\multicolumn{5}{c}{{SFSC}} \\

 \midrule
    $\phi_0$& 69.51 & 69.30& 69.07 & 68.66 & 67.77 & 72.36 & 57.82 & 55.61 & 55.05 & 52.32& 71.03 & 71.01 & 70.90 & 70.51 & 69.46\\
    $\phi_{80\%}$&69.37  &69.14 & 68.91 & 68.55 & 67.66 &56.49 & 52.21 &47.94& 47.13& 43.39 & 71.19 & 71.19 & 71.06 & 70.67 & 69.62\\
    $\phi_{60\%}$&69.17  & 68.96 &68.77  & 68.43 & 67.53 &  55.50& 48.80 & 49.57  & 47.08 & 43.46& 71.09 & 71.08 & 71.03 & 70.65 & 69.53 \\
    $\phi_{40\%}$&  68.92&68.71& 68.45 & 68.21 & 67.44 &  54.42& 47.98 & 46.83 &48.80 & 44.52 & 70.18 & 70.16 & 70.15 & 69.81 & 68.64\\
    $\phi_{20\%}$& 68.20 & 68.00 & 67.82 &67.56 & 66.92 & 52.62 & 45.15 & 44.36 &45.71& 45.64& 69.58 & 69.62 & 69.55 & 69.23 & 68.17 \\
\midrule
\multicolumn{1}{c}{}&\multicolumn{5}{c}{BCT-S w/ PrunNet} &\multicolumn{5}{c}{Asymmetric-S w/ PrunNet}&\multicolumn{5}{c}{{\textbf{Ours}}}\\
 \midrule
     $\phi_0$& 69.98 & 69.98 & 69.98 & 69.98 & 69.90 &  72.36 & 72.36 & 72.52& 71.56 & 69.94 & \textbf{74.60} & \textbf{74.59} &\textbf{74.57} & \textbf{74.53} & \textbf{74.38}\\
    $\phi_{80\%}$&69.89 & 70.02 & 69.98 & 69.98 & 69.89 &72.34 &72.36 & 72.50& 71.55 & 69.97& \textbf{74.62} & \textbf{74.62} & \textbf{74.60} & \textbf{74.55} & \textbf{74.40}\\
    $\phi_{60\%}$&69.98 & 69.98 &70.01 &69.98 & 69.90 & 72.17 & 72.16 & 72.29 & 71.37 & 69.70& \textbf{74.65} & \textbf{74.64} & \textbf{74.61} & \textbf{74.58} & \textbf{74.44} \\
    $\phi_{40\%}$&70.01 & 70.01& 70.01 & 70.02 & 69.94 & 71.27 & 71.26 & 71.36 & 70.53 & 68.99& \textbf{74.53} & \textbf{74.52} &\textbf{74.50}& \textbf{74.47} & \textbf{74.31}\\
    $\phi_{20\%}$&69.94 & 69.94 & 69.94& 69.94& 69.88 & 70.00 & 70.01 & 70.07 & 69.33& 68.51& \textbf{74.35} & \textbf{74.35} & \textbf{74.31} &\textbf{74.28} & \textbf{74.18} \\
\midrule
\rowcolor[gray]{0.9}\multicolumn{16}{c}{\emph{ROXford}}\\
\midrule

\multicolumn{1}{c}{}&\multicolumn{5}{c}{Independent learning} &\multicolumn{5}{c}{Joint learning} &\multicolumn{5}{c}{O2O-SSPL}\\
\midrule
    $\phi_0$& 52.28 & -- & -- & -- & -- & 50.23 & 50.02 & 50.28 & 48.34 & 48.67 & 52.28 &49.24& 49.48 & 48.61 & 46.29 \\
    $\phi_{80\%}$& -- & 51.94& -- & -- & --  & 48.57 & 49.47 & 49.28 & 47.46 & 48.03 & 50.51 &  46.20 & 47.49 & 46.73 & 44.40 \\
    $\phi_{60\%}$& -- & -- & 51.00 & -- & -- & 48.97 & 49.51 & 50.17 & 47.90 & 48.61 & 50.15 & 45.67 & 47.44 & 46.10 & 43.82 \\
    $\phi_{40\%}$& -- & -- & -- & 50.26 & -- & 48.15 & 49.10 & 49.70 & 48.69 & 47.49 & 49.64 &46.71 &  46.85 & 46.34& 43.93 \\
    $\phi_{20\%}$& -- & -- & -- & -- & 49.32 & 48.20 & 48.59 & 49.90 & 46.76 & 48.30 & 48.82 & 44.66 &46.14 & 44.48 &  43.98 \\
\midrule
\multicolumn{1}{c}{}&\multicolumn{5}{c}{{BCT-S w/ SwitchNet}} &\multicolumn{5}{c}{Asymmetric-S w/ SwitchNet}&\multicolumn{5}{c}{{SFSC}} \\
\midrule
    $\phi_0$&52.51  & \textbf{52.75} & 52.02& 50.89 & 48.97 & 51.90& 36.31 & 36.37 & 36.97 & 32.67& 52.59 & 52.40 & 51.71 & 50.86 & 49.35 \\
    $\phi_{80\%}$& 52.49 &\textbf{53.51} & \textbf{52.98} & 50.66 & 48.79& 40.36 & 34.69 & 29.60 & 30.26 & 26.99& 52.31 & 51.96 & 51.37 & 50.90 & 49.30 \\
    $\phi_{60\%}$& \textbf{52.60} & \textbf{53.16} & \textbf{52.82} & 50.22 &48.11 &39.62 & 31.46 & 32.80  & 30.40 & 27.75 & 51.24 & 51.77 & 51.67 & 51.17 & 49.65 \\
    $\phi_{40\%}$& 51.47 & 51.46 & 51.46 & 51.08 & 51.28 &37.98  & 30.83 & 31.05 & 33.28 & 28.22 & 51.74 & 51.48& 51.22 & 50.36 & 48.83 \\
    $\phi_{20\%}$& 50.67 & 50.57 & 50.73 & 49.00& 46.97 & 37.30 & 28.28 & 29.83 &  31.54 & 29.60  & 50.98 & 50.94& 50.44 & 49.77 & 48.12 \\
\midrule
\multicolumn{1}{c}{}&\multicolumn{5}{c}{BCT-S w/ PrunNet} &\multicolumn{5}{c}{Asymmetric-S w/ PrunNet}&\multicolumn{5}{c}{{\textbf{Ours}}}\\
\midrule
    $\phi_0$&51.54 & 51.53 & 51.54 & 51.14 & 51.31 & 51.80 & 51.88 & 51.60& 51.29 & 49.28&\textbf{52.69}& 52.68 & \textbf{52.73} & \textbf{52.68} & \textbf{52.38}\\
    $\phi_{80\%}$&51.54 & 51.54 & 51.57 & 51.14 & 51.30 &52.25 &52.33 & 52.12 & 51.43 &49.21&\textbf{52.67} & 52.66 & 52.64 &\textbf{52.64} & \textbf{52.38}\\
    $\phi_{60\%}$& 51.55 & 51.54 & 51.51 & 51.13 & 51.28& 51.21 & 51.28 & 51.35 & 51.07 & 48.66 & 52.59 &52.61 & 52.59 & \textbf{52.65} & \textbf{52.43}\\
    $\phi_{40\%}$& 51.47 & 51.46 & 51.46 & 51.08 & 51.28 & 50.78 & 50.80 & 50.83 & 49.64 & 48.43& \textbf{51.99} &\textbf{51.99}&\textbf{51.91} & \textbf{51.95} & \textbf{51.76}\\
    $\phi_{20\%}$&\textbf{51.27} & \textbf{51.28} & 51.26 & 51.13 & 51.29 & 49.52 & 50.10 & 49.78 &49.40 & 47.34& 51.19 & 51.22 & \textbf{51.27} &\textbf{51.63}& \textbf{51.49}\\
\midrule
\rowcolor[gray]{0.9}\multicolumn{16}{c}{\emph{GLDv2-test}}\\
\midrule

\multicolumn{1}{c}{}&\multicolumn{5}{c}{Independent learning} &\multicolumn{5}{c}{Joint learning} &\multicolumn{5}{c}{O2O-SSPL}\\
 \midrule
    $\phi_0$& 10.59 & -- & -- & -- & -- &  10.02 & 9.70 & 9.31&8.85& 8.92&10.59 & 9.92 &10.00 & 9.86 & 9.23 \\
    $\phi_{80\%}$& -- & 10.39& -- & -- & --  & 9.72 & 9.95 &9.30 & 8.82 &9.01 &9.94 & 9.60 & 9.62 & 9.47& 8.84  \\
    $\phi_{60\%}$& -- & -- & 9.94 & -- & -- & 9.55 & 9.54 & 9.59 & 8.85 & 8.98& 9.72& 9.30 & 9.58 & 9.30 & 8.67  \\
    $\phi_{40\%}$& -- & -- & -- & 9.58 & -- & 9.28 &9.00 & 8.72 &8.74  & 8.47&9.29 &8.96 & 8.97 & 9.07 & 8.36  \\
    $\phi_{20\%}$& -- & -- & -- & -- & 8.23 & 8.74 & 8.71 &8.13 &8.01 & 8.47& 8.77& 8.55 & 8.61 & 8.64 & 8.38  \\
\midrule
\multicolumn{1}{c}{}&\multicolumn{5}{c}{{BCT-S w/ SwitchNet}} &\multicolumn{5}{c}{Asymmetric-S w/ SwitchNet}&\multicolumn{5}{c}{{SFSC}} \\
\midrule
    $\phi_0$& 9.29 & 9.20& 9.03 & 8.85 &8.58 & 11.00 & 5.21 & 5.01& 5.10 & 3.64&9.79&9.75&9.43&9.25&8.54 \\
    $\phi_{80\%}$&  9.22&9.19 & 9.04 & 8.84 & 8.55 & 4.32& 4.26 & 3.20 & 2.87 & 2.11&9.70&9.69&9.28&9.08&8.50\\
    $\phi_{60\%}$& 9.08 & 9.03&  8.99&8.79  & 8.53 & 3.86 & 2.96 & 3.87  & 2.93 & 2.00&9.48&9.38&9.04&8.91&8.38 \\
    $\phi_{40\%}$& 8.84 & 8.79& 8.75 & 8.74 &8.43  & 3.48 & 2.59 & 2.74 & 3.42 & 2.10&9.08&9.07&8.80&8.78 &8.30\\
    $\phi_{20\%}$& 8.12 & 8.15 & 8.11 & 8.10 &  8.22& 2.63 & 2.06 & 2.07 & 2.42& 2.55&8.45&8.35&8.14&8.05&8.00 \\
\midrule
\multicolumn{1}{c}{}&\multicolumn{5}{c}{BCT-S w/ PrunNet} &\multicolumn{5}{c}{Asymmetric-S w/ PrunNet}&\multicolumn{5}{c}{{\textbf{Ours}}}\\
\midrule
    $\phi_0$&  9.59 & 9.60 & 9.63 & 9.62 & 9.56& 11.36 & 11.38 & 11.15 & 10.72 & 9.61& \textbf{11.59} & \textbf{11.60} &\textbf{11.60}& \textbf{11.60} & \textbf{11.48} \\
    $\phi_{80\%}$& 9.61&9.61  & 9.63 & 9.62 & 9.56 &11.32 &11.51 & 11.18 & 10.72 & 9.63& \textbf{11.57} & \textbf{11.59} & \textbf{11.60}& \textbf{11.59} & \textbf{11.44}\\
    $\phi_{60\%}$& 9.60 & 9.61 & 9.62 & 9.63 & 9.59 & 11.13 & 11.23 & 11.01 & 10.71 & 9.55& \textbf{11.54} & \textbf{11.54} & \textbf{11.56} & \textbf{11.55} & \textbf{11.37}\\
    $\phi_{40\%}$& 9.64 & 9.65 & 9.65 & 9.67 & 9.59 & 10.34 & 10.38 & 10.57 & 10.40 & 9.39& \textbf{11.41} & \textbf{11.45} & \textbf{11.43} & \textbf{11.49} & \textbf{11.38}\\
    $\phi_{20\%}$& 9.55 & 9.55 & 9.55 &9.57 & 9.53 & 9.02 & 9.20 & 9.30 &9.19& 8.89& \textbf{11.30} & \textbf{11.32} & \textbf{11.30} & \textbf{11.30} & \textbf{11.22}\\
\bottomrule
\end{tabular}}
\label{Tab:performance_landmark_supple}
\end{table*}

\begin{table*}[t!]
\centering
\setlength{\tabcolsep}{3pt}
\renewcommand{\arraystretch}{1.05}
\caption{Detailed comparisons on pre-determined capacities over RParis~\cite{rparis&roxford}, ROxford~\cite{rparis&roxford} and GLDv2-test~\cite{gldv2} using different backbones.}
\resizebox{1.0\linewidth}{!}{
\begin{tabular}{ll|c|cc|cc|cc|cc}
\toprule
&& $\mathcal{M}(\phi_0, \phi_0)$ & $\mathcal{M}(\phi_{80\%}, \phi_{80\%})$ & $\mathcal{M}(\phi_{80\%}, \phi_0)$ & $\mathcal{M}(\phi_{60\%}, \phi_{60\%})$ & $\mathcal{M}(\phi_{60\%}, \phi_0)$ & $\mathcal{M}(\phi_{40\%}, \phi_{40\%})$ & $\mathcal{M}(\phi_{40\%}, \phi_0)$ & $\mathcal{M}(\phi_{20\%}, \phi_{20\%})$ & $\mathcal{M}(\phi_{20\%}, \phi_0)$\\
&& Self-test & Self-test & Cross-test & Self-test & Cross-test & Self-test & Cross-test & Self-test & Cross-test\\
\midrule
\rowcolor[gray]{0.9}\multicolumn{11}{c}{\emph{RParis}} \\
\midrule
    \multirow{3}{*}{ResNet-50}& Independent learning & 74.33  & 73.94 & -- & 73.75 & -- & 73.44 & -- & 72.82 & --\\
    &SFSC &74.59&74.48&74.52&74.32&74.43&74.25&74.36&73.61&74.04\\
    &\textbf{Ours}& \textbf{75.05}  & \textbf{75.01} & \textbf{75.02} &\textbf{74.95} & \textbf{74.96} & \textbf{74.90} & \textbf{74.91} & \textbf{74.78}&\textbf{74.93} \\
\midrule
    \multirow{3}{*}{ResNeXt-50}& Independent learning & 75.22&75.03 & -- & 74.63 & -- & 73.77 & -- & 70.71&-- \\
    &SFSC &74.92&73.80&73.78&73.67&73.71&72.50&73.16&71.22&73.08\\
    &\textbf{Ours} & \textbf{76.03}  &\textbf{75.97}&\textbf{75.97}&\textbf{75.94} & \textbf{75.90}&\textbf{75.77}& \textbf{75.80} & \textbf{75.07}&\textbf{75.36 }\\
\midrule
    \multirow{3}{*}{MobileNet-V2}&Independent learning& 66.60& 65.76 & -- & 65.05 & -- & 64.51 & -- & 63.68&-- \\
    &SFSC &66.38&65.91&66.10&65.75&66.08&65.27&65.81&63.83&65.09\\
    &\textbf{Ours} &\textbf{67.15}  &\textbf{67.10} & \textbf{67.08} & \textbf{66.95}& \textbf{67.05} &\textbf{66.53} &\textbf{66.84}& \textbf{64.57} &\textbf{66.01}\\
\midrule
    \multirow{3}{*}{ViT-Small}& Independent learning & 80.81 & 73.40 &--   & 70.87 & -- & 64.61 & -- & 52.93&-- \\
    &SFSC &77.37&74.42&75.28&70.72&73.02&68.66&72.76&55.15&63.83\\
    &\textbf{Ours} & \textbf{82.00}  & \textbf{80.99} & \textbf{81.22}& \textbf{80.54} & \textbf{80.72}& \textbf{77.74} & \textbf{78.73} &\textbf{72.22}  & \textbf{74.24} \\
\midrule
\rowcolor[gray]{0.9}\multicolumn{11}{c}{\emph{ROxford}} \\
\midrule
    \multirow{3}{*}{ResNet-50}& Independent learning & 54.70  & 54.56 & -- & 54.14 & -- & 54.20 & -- & 50.90 & --\\
    &SFSC  &53.84&53.75&53.73&53.35&53.62&53.22&53.26&52.88&53.19\\
    &\textbf{Ours}& \textbf{56.12}  & \textbf{55.81} & \textbf{55.97} &\textbf{55.71} & \textbf{55.98} & \textbf{55.38} & \textbf{55.84} & \textbf{54.69}&\textbf{55.52} \\
\midrule
    \multirow{3}{*}{ResNeXt-50}& Independent learning & 55.38&53.73 & -- & 52.61 & -- & 52.16 & -- & 50.87&-- \\
    &SFSC  & 54.57&54.06&53.52&53.06&53.09&52.40&53.21&50.31&52.40\\
    &\textbf{Ours} & \textbf{57.63}  &\textbf{57.73}&\textbf{57.75}&\textbf{57.82} & \textbf{57.76}&\textbf{58.27}& \textbf{57.96} & \textbf{56.73}&\textbf{57.45} \\
\midrule
    \multirow{3}{*}{MobileNet-V2}&Independent learning& 46.60& 45.91 & -- & 45.62 & -- & 44.39 & -- & 43.88&-- \\
    &SFSC & 46.84&45.64&45.07&46.44&46.56&45.51&46.07&43.22&43.90\\
    &\textbf{Ours} &\textbf{47.63}  &\textbf{47.88} & \textbf{47.64} & \textbf{48.09}& \textbf{47.83} &\textbf{47.17} &\textbf{47.55}& \textbf{45.20} &\textbf{46.71}\\
\midrule
    \multirow{3}{*}{ViT-Small}& Independent learning & 59.88 & 54.45 &--   & 46.41 & -- &37.22 & -- &28.58 &-- \\
    &SFSC & 56.10&52.25&54.98&48.24&54.68&43.60&48.68&31.05&37.85\\
    &\textbf{Ours} & \textbf{60.11}  &\textbf{55.36} &\textbf{55.46} & \textbf{54.84} &\textbf{56.01} & \textbf{52.24}&\textbf{52.70} & \textbf{43.96} &\textbf{46.50}  \\
\midrule
\rowcolor[gray]{0.9}\multicolumn{11}{c}{\emph{GLDv2-test}} \\
\midrule
    \multirow{3}{*}{ResNet-50}& Independent learning & 12.15  & 12.03 & -- & 11.52 & -- & 11.38 & -- & 10.50 & --\\
    &SFSC & 10.84&10.89&11.01&10.91&11.02&10.86&10.90&9.96&10.30\\
    &\textbf{Ours}& \textbf{12.46}  & \textbf{12.41} & \textbf{12.38} &\textbf{12.49} & \textbf{12.43} & \textbf{12.46} & \textbf{12.45} & \textbf{12.18}&\textbf{12.39} \\
\midrule
    \multirow{3}{*}{ResNeXt-50}& Independent learning & 12.92&11.95 & -- & 11.54 & -- & 11.41 & -- & 10.05&-- \\
    &SFSC  &11.77&11.23&11.42&10.83&11.14&10.43&10.86&9.19&9.99 \\
    &\textbf{Ours} & \textbf{13.03}  &\textbf{13.05}&\textbf{13.01}&\textbf{13.11} & \textbf{13.02}&\textbf{12.98}& \textbf{12.99} & \textbf{12.84}&\textbf{12.92} \\
\midrule
    \multirow{3}{*}{MobileNet-V2}&Independent learning& 8.38& 7.94 & -- & 7.65 & -- & 6.65 & -- & 6.30&-- \\
    &SFSC & 7.50&7.42&7.56&7.31&7.26&6.78&7.03&6.23&6.48\\
    &\textbf{Ours} &\textbf{8.80} &\textbf{8.82} & \textbf{8.82} & \textbf{8.83}& \textbf{8.78} &\textbf{8.47} &\textbf{8.70}& \textbf{7.13} &\textbf{7.77}\\
\midrule
    \multirow{3}{*}{ViT-Small}& Independent learning & 15.03 & 12.28 &--   & 8.89 & -- &5.43 & -- &2.96 &-- \\
    &SFSC &13.40&11.01&11.90&9.71&10.68&6.89&8.19&3.33&4.35\\
    &\textbf{Ours} & \textbf{15.06} &\textbf{13.96} & \textbf{14.26}& \textbf{14.01} & \textbf{14.29}& \textbf{12.18}& \textbf{12.71}& \textbf{7.45} &\textbf{9.47}  \\
\bottomrule
\end{tabular}}
\label{Tab: different model architecture supplement}
\end{table*}

\begin{table*}[t!]
\centering
\setlength{\tabcolsep}{10pt}
\renewcommand{\arraystretch}{1.0}
\caption{Detailed comparisons on the new capacity (10\%) over RParis~\cite{rparis&roxford}, ROxford~\cite{rparis&roxford}, and GLDv2-test~\cite{gldv2}. ResNet-18 is used as the backbone. For methods without PrunNet, we use BCT~\cite{BCT} or SSPL~\cite{wu2023_SSPL} to train a new small-capacity model, whose capacity is 10\% of the dense network $\phi_0$, with compatibility with existing models.}
\resizebox{1.0\linewidth}{!}{
\begin{tabular}{lcccccc}
\toprule
Methods & $\mathcal{M}(\phi_{10\%}, \phi_0)$ & $\mathcal{M}(\phi_{10\%}, \phi_{80\%})$ & $\mathcal{M}(\phi_{10\%}, \phi_{60\%})$& $\mathcal{M}(\phi_{10\%}, \phi_{40\%})$ & $\mathcal{M}(\phi_{10\%}, \phi_{20\%})$ & $\mathcal{M}(\phi_{10\%}$, $\phi_{10\%})$ \\
\midrule
\rowcolor[gray]{0.9}\multicolumn{7}{c}{\emph{RParis}}\\
\midrule
Joint learning + BCT & 70.27 & 69.84 & 69.62 & 68.87 & 68.48 & 68.25\\
O2O-SSPL +SSPL & 68.83 &67.62&67.09&66.90&65.33&64.30 \\
BCT-S w/ SwitchNet + BCT  &67.99&67.82&67.61&67.18&66.57&66.07\\
Asymmetric-S w/ SwitchNet + BCT  &68.86&56.88&54.66&54.10&51.29&67.07 \\
SFSC + BCT  &68.71&68.53&68.61&68.32&67.43&66.71\\
BCT-S w/ PrunNet &68.79&68.79&68.81&68.80&68.67&65.93\\
Asymmetric-S w/ PrunNet &62.27&62.21&62.37&61.94&61.63 &56.09\\
\textbf{Ours} &\textbf{73.42}&\textbf{73.41}&\textbf{73.40}&\textbf{73.41}&\textbf{73.38}&\textbf{70.12}\\
\midrule
\rowcolor[gray]{0.9}\multicolumn{7}{c}{\emph{ROxford}} \\
\midrule
Joint learning + BCT &\textbf{50.65}&\textbf{50.55}&48.73&49.03&48.68&47.48 \\
O2O-SSPL + SSPL &47.95&45.24&45.82&44.92&42.10&43.05\\
BCT-S w/ SwitchNet +BCT &49.39&49.38&49.10&47.72&45.36&46.22\\
Asymmetric-S w/ SwitchNet + BCT &50.38&34.81&35.05&35.08&31.37&47.71\\
SFSC + BCT &48.84&48.95&48.57&47.40&45.09&44.49\\
BCT-S w/ PrunNet &49.30&49.28&49.27&49.20&49.18&47.03\\
Asymmetric-S w/ PrunNet &46.27&46.74&46.48&46.84&44.88&42.48\\
\textbf{Ours}&50.53&50.42&\textbf{50.54}&\textbf{50.67}&\textbf{50.40}&\textbf{48.41}\\
\midrule
\rowcolor[gray]{0.9}\multicolumn{7}{c}{\emph{GLDv2-test}} \\
\midrule
Joint learning + BCT &8.47&8.78&8.39&8.36&8.17&8.11 \\
O2O-SSPL + SSPL &7.68&7.69&7.82&7.62&7.29&6.95\\
BCT-S w/ SwitchNet + BCT &7.75&7.72&7.68&7.47&7.41&7.73\\
Asymmetric-S w/SwitchNet + BCT &8.48&4.40&4.05&4.39&3.19&8.27\\
SFSC + BCT &7.23&7.14&6.93&7.03&6.84&7.47\\
BCT-S w/ PrunNet &8.21&8.19&8.20&8.23&8.28&8.01\\
Asymmetric-S w/ PrunNet&4.20&4.24&4.34&4.30&4.62&4.09 \\
\textbf{Ours}&\textbf{10.07}&\textbf{10.05}&\textbf{10.04}&\textbf{10.08}&\textbf{9.87}&\textbf{9.12}\\
\bottomrule
\end{tabular}}
\label{Tab:new capacity supple}
\end{table*}

\begin{table*}[t!]
\centering
\setlength{\tabcolsep}{6pt}
\renewcommand{\arraystretch}{1.05}
\caption{Detailed results of different variants over RParis~\cite{rparis&roxford}, ROxford~\cite{rparis&roxford} and GLDv2-test~\cite{gldv2}. ResNet-18 is used as the backbone.}
\resizebox{1.0\linewidth}{!}{
\begin{tabular}{l|c|cc|cc|cc|cc}
\toprule
& $\mathcal{M}(\phi_0, \phi_0)$ & $\mathcal{M}(\phi_{80\%}, \phi_{80\%})$ & $\mathcal{M}(\phi_{80\%}, \phi_0)$ & $\mathcal{M}(\phi_{60\%}, \phi_{60\%})$ & $\mathcal{M}(\phi_{60\%}, \phi_0)$ & $\mathcal{M}(\phi_{40\%}, \phi_{40\%})$ & $\mathcal{M}(\phi_{40\%}, \phi_0)$ & $\mathcal{M}(\phi_{20\%}, \phi_{20\%})$ & $\mathcal{M}(\phi_{20\%}, \phi_0)$\\
& Self-test & Self-test & Cross-test & Self-test & Cross-test & Self-test & Cross-test & Self-test & Cross-test\\
\midrule
\rowcolor[gray]{0.9}\multicolumn{10}{c}{\emph{RParis}} \\
\midrule
    Independent learning & 73.35 & 71.84 & --&70.71&--&70.37&--&67.77&-- \\
    \textbf{Ours} ($N$ = 4) & \textbf{74.60} & \textbf{74.62} & \textbf{74.62} & \textbf{74.61}&\textbf{74.65}&\textbf{74.47}&\textbf{74.53}&\textbf{74.18}&\textbf{74.35} \\
    Frozen scores &72.72&72.66&72.72&72.61&72.76&72.18&72.72&69.72&71.37\\
    $N$ score maps &72.01&71.45&71.57&71.38&71.69&70.88&71.38&69.57&70.99\\
    \midrule
    Direct gradient integration& 73.09&73.06&73.09&73.07&73.08&73.90&73.10&72.64&72.79\\
    Direct loss combination &69.51&69.14&69.37&68.77&69.17&68.21&68.92&66.92&68.20\\
    Pareto integration &72.10&72.09&72.09&72.11&72.09&72.04&72.10&71.36&71.71\\
    \midrule
    \textbf{Ours} ($N$ = 1)&72.33  & 72.36 &72.33&72.35&72.32&72.21&72.23&67.11&70.28 \\
    \textbf{Ours} ($N$ = 2)&73.56  & 73.45 & 73.49&73.39&73.41&73.31&73.42&70.32&72.31 \\
    \textbf{Ours} ($N$ = 6)&73.99 & 74.02 & 74.01&73.98&73.99&73.65&73.81&72.96&73.33\\
    \textbf{Ours} ($N$ = 8)&73.58 & 73.56 & 73.58&73.52&73.56&73.47&73.54&72.76&73.14\\
\midrule
\rowcolor[gray]{0.9}\multicolumn{10}{c}{\emph{ROxford}} \\
\midrule
    Independently learning & 52.28 & 51.94 & --&51.00&--&50.26&--&49.32&-- \\
    \textbf{Ours} ($N$=4) &52.69 &52.66&52.67&52.59&52.59&51.95&51.99&51.49&51.19\\
    Frozen scores & 52.03&51.86&51.95&51.74&51.80&51.78&51.87&50.08&50.80\\
    $N$ score maps &50.11&49.87&49.56&49.37&48.79&48.73&49.41&48.02&48.73\\
    \midrule
    Direct gradient integration & 52.53&52.49&52.50&52.22&52.48&52.12&52.49&52.04&52.25 \\
    Direct loss combination & 51.54&51.54&51.54&51.54&51.55&51.46&51.47&51.29&51.27\\
    Pareto integration &51.85&51.73&51.84&51.70&51.82&51.44&51.73&49.97&51.38\\
    \midrule
    \textbf{Ours} ($N$ = 1) &51.20&51.36&51.02&51.16&50.88&51.22&51.29&46.37&46.81\\
    \textbf{Ours} ($N$ = 2) &52.00&51.87&51.87&51.87&51.94&51.70&51.93&47.43&51.24\\
    \textbf{Ours} ($N$ = 6)& \textbf{53.82}&\textbf{53.75}&\textbf{53.76}&\textbf{53.67}&\textbf{53.74}&\textbf{53.42}&\textbf{53.62}&\textbf{52.32}&\textbf{53.37}\\
    \textbf{Ours} ($N$ = 8) &52.63&52.58&52.60&52.68&52.66&52.83&52.82&52.14&52.93\\
\midrule
\rowcolor[gray]{0.9}\multicolumn{10}{c}{\emph{GLDv2-test}} \\
\midrule
    Independently learning & 10.59 &10.39 & --&9.94&--&9.58&--&8.23 \\
    \textbf{Ours} ($N$=4) &\textbf{11.59}&\textbf{11.59}&\textbf{11.57}&\textbf{11.56}&\textbf{11.54}&\textbf{11.49}&\textbf{11.41}&\textbf{11.22}&\textbf{11.30}\\
    Frozen scores &10.95&10.81&10.86&10.69&10.71&10.12&10.42&9.21&9.71\\
    $N$ score maps&10.66&10.57&10.59&10.18&10.39&10.06&10.37&9.11&9.43\\
    \midrule
    Direct gradient integration &11.48&11.47&11.45&11.47&11.47&11.35&11.36&11.21&11.28\\
    Direct loss combination &9.59&9.61&9.61&8.99&9.08&8.74&8.84&8.22&8.12\\
    Pareto integration &10.57&10.57&10.58&10.58&10.58&10.62&10.63&10.23&10.39\\
    \midrule
    \textbf{Ours} ($N$ = 1) &11.03&11.06&11.06&10.95&11.03&10.77&10.85&9.07&9.55\\
    \textbf{Ours} ($N$ = 2) &11.33&11.37&11.31&11.39&11.36&11.16&11.24&9.46&10.22\\
    \textbf{Ours} ($N$ = 6)& 11.18&11.17&11.17&11.16&11.16&11.01&11.10&10.91&11.07\\
    \textbf{Ours} ($N$ = 8) &11.45&11.47&11.47&11.40&11.41&11.39&11.36&11.01&11.02\\
\bottomrule
\end{tabular}}
\label{Tab:ablation_suppl}
\end{table*}

\begin{table*}[h!]
\centering
\setlength{\tabcolsep}{6pt}
\renewcommand{\arraystretch}{1.0}
\caption{Comparisons on pre-determined capacities over Market-1501~\cite{MARKET1501}. We employ ResNet-18 as the backbone. We use the same setting for the subnetwork capacities as SFSC~\cite{wu2023_SFSC} to include the results reported by~\cite{wu2023_SFSC} (denoted by $\dagger$) in the comparison on Market-1501.}
\vspace{-2mm}
\resizebox{1.0\linewidth}{!}{
\begin{tabular}{l|c|cc|cc|cc}
\toprule
 & $\mathcal{M}(\phi_0,\phi_0)$ & $\mathcal{M}(\phi_{56.25\%},\phi_{56.25\%})$ & $\mathcal{M}(\phi_{56.25\%},\phi_0)$ & $\mathcal{M}(\phi_{25\%},\phi_{25\%})$ & $\mathcal{M}(\phi_{25\%},\phi_0)$ & $\mathcal{M}(\phi_{6.25\%},\phi_{6.25\%})$ & $\mathcal{M}(\phi_{6.25\%},\phi_0)$ \\
 & Self-test & Self-test & Cross-test & Self-test & Cross-test & Self-test & Cross-test \\
\midrule
    Independent learning &80.91&71.25&--&67.48&--&55.25&-- \\
    SFSC$^{\dagger}$ & 81.43&72.06&77.26&70.74&76.37&58.19&69.43\\
    \textbf{Ours} &\textbf{81.55}& \textbf{81.25} &\textbf{81.36}&\textbf{81.32} &\textbf{81.28}&\textbf{80.08}&\textbf{80.31}\\
\bottomrule
\end{tabular}}
\label{Tab: market1501 results}
\end{table*}

\begin{table*}[h!]
\centering
\setlength{\tabcolsep}{6pt}
\renewcommand{\arraystretch}{1.0}
\caption{Comparisons on pre-determined capacities over MSMT17~\cite{MSMT17}. We employ ResNet-18 as the backbone. We use the same setting for the subnetwork capacities as SFSC~\cite{wu2023_SFSC} to include the results reported by~\cite{wu2023_SFSC} (denoted by $\dagger$) in the comparison on MSMT17.}
\vspace{-2mm}
\resizebox{1.0\linewidth}{!}{
\begin{tabular}{l|c|cc|cc|cc}
\toprule
 & $\mathcal{M}(\phi_0,\phi_0)$ & $\mathcal{M}(\phi_{56.25\%},\phi_{56.25\%})$ & $\mathcal{M}(\phi_{56.25\%},\phi_0)$ & $\mathcal{M}(\phi_{25\%},\phi_{25\%})$ & $\mathcal{M}(\phi_{25\%},\phi_0)$ & $\mathcal{M}(\phi_{6.25\%},\phi_{6.25\%})$ & $\mathcal{M}(\phi_{6.25\%},\phi_0)$ \\
 & Self-test & Self-test & Cross-test & Self-test & Cross-test & Self-test & Cross-test \\
\midrule
    Independent learning &43.06&30.06&--&22.86&--&11.69&-- \\
    SFSC$^{\dagger}$ &43.89&--&37.74&--&35.32&--&28.16\\
    \textbf{Ours} &\textbf{44.73}& \textbf{43.93} &\textbf{44.26}&\textbf{42.77} &\textbf{43.58}&\textbf{41.29}&\textbf{42.75}\\
\bottomrule
\end{tabular}}
\label{Tab: MSMT17 results}
\end{table*}

\begin{table*}[h]
\centering
\setlength{\tabcolsep}{5pt}
\renewcommand{\arraystretch}{1.0}
\caption{Recall@1 on CUB-200~\cite{CUB200}. We employ ViT-S as the backbone. All models are pretrained on ImageNet-1k before being fine-tuned on CUB-200.}
\resizebox{1.0\linewidth}{!}{
\begin{tabular}{l|c|cc|cc|cc|cc}
\toprule
 & $\mathcal{M}(\phi_0,\phi_0)$ & $\mathcal{M}(\phi_{80\%},\phi_{80\%})$ &$\mathcal{M}(\phi_{80\%},\phi_{0})$ & $\mathcal{M}(\phi_{60\%},\phi_{60\%})$ & $\mathcal{M}(\phi_{60\%},\phi_{0})$ & $\mathcal{M}(\phi_{40\%},\phi_{40\%})$ & $\mathcal{M}(\phi_{40\%},\phi_{0})$ & $\mathcal{M}(\phi_{20\%},\phi_{20\%})$ &$\mathcal{M}(\phi_{20\%},\phi_{0})$\\
 & Self-test & Self-test & Cross-test & Self-test & Cross-test & Self-test & Cross-test & Self-test & Cross-test\\
\midrule
     Independent learning &80.00 &78.89&--&78.43&--&78.18&--& 77.61&--  \\
     SFSC &80.54&80.43&80.55&80.27&80.41&80.15&80.24&78.79&78.68  \\
     \textbf{Ours} &\textbf{82.46} & \textbf{82.45} & \textbf{82.53} & \textbf{82.29} & \textbf{82.41} & \textbf{81.57} & \textbf{81.72} & \textbf{79.32} & \textbf{79.91}  \\
\bottomrule
\end{tabular}}
\label{Tab: CUB200 results}
\vspace{-3mm}
\end{table*}

\begin{table*}[h]
\centering
\setlength{\tabcolsep}{5pt}
\renewcommand{\arraystretch}{1.0}
\caption{Comparison on PrunNet implemented by structured pruning (Str.) and unstructured pruning (UnStr.) on Landmark datasets (Average mAP) and Inshop dataset (Recall@1). We employ ResNet-18 as the backbone.}
\resizebox{1.0\linewidth}{!}{
\begin{tabular}{ll|c|cc|cc|cc|cc}
\toprule
&& $\mathcal{M}(\phi_0,\phi_0)$ & $\mathcal{M}(\phi_{80\%},\phi_{80\%})$ &$\mathcal{M}(\phi_{80\%},\phi_{0})$ & $\mathcal{M}(\phi_{60\%},\phi_{60\%})$ & $\mathcal{M}(\phi_{60\%},\phi_{0})$ & $\mathcal{M}(\phi_{40\%},\phi_{40\%})$ & $\mathcal{M}(\phi_{40\%},\phi_{0})$ & $\mathcal{M}(\phi_{20\%},\phi_{20\%})$ &$\mathcal{M}(\phi_{20\%},\phi_{0})$\\
 && Self-test & Self-test & Cross-test & Self-test & Cross-test & Self-test & Cross-test & Self-test & Cross-test\\
\midrule
\multirow{3}{*}{Landmark} & SFSC & 44.47 & 44.28 & 44.40 & 43.91 & 43.94 & 42.98 & 43.67 & 41.43 & 43.00 \\
    & \textbf{Ours} (Str.) &44.81&	44.72	&45.04	&44.46	&45.14&	44.07&	44.33&	41.58&	43.10\\
    & \textbf{Ours} (UnStr.) & \textbf{46.29} & \textbf{46.29} & \textbf{46.29} & \textbf{46.25}&\textbf{46.26}&\textbf{45.97}&\textbf{45.98}&\textbf{45.63}&\textbf{45.61} \\
\midrule
\multirow{3}{*}{In-shop} & SFSC & 84.57  & 84.48 & 84.40&84.25&84.31&84.15&84.20&83.57&83.74 \\
    & \textbf{Ours} (Str.)& 86.90&86.69& 86.78&86.59&86.70&86.37&86.66& 86.19 &86.34\\
    & \textbf{Ours} (UnStr.) & \textbf{87.31} & \textbf{87.30} & \textbf{87.33} & \textbf{87.21}&\textbf{87.23}&\textbf{87.14}&\textbf{87.15}&\textbf{86.43}&\textbf{86.77} \\
\bottomrule
\end{tabular}}
\label{Tab: structured}
\vspace{-3mm}
\end{table*}

\begin{figure*}[t!]
\centering
\includegraphics[width=0.95\linewidth]{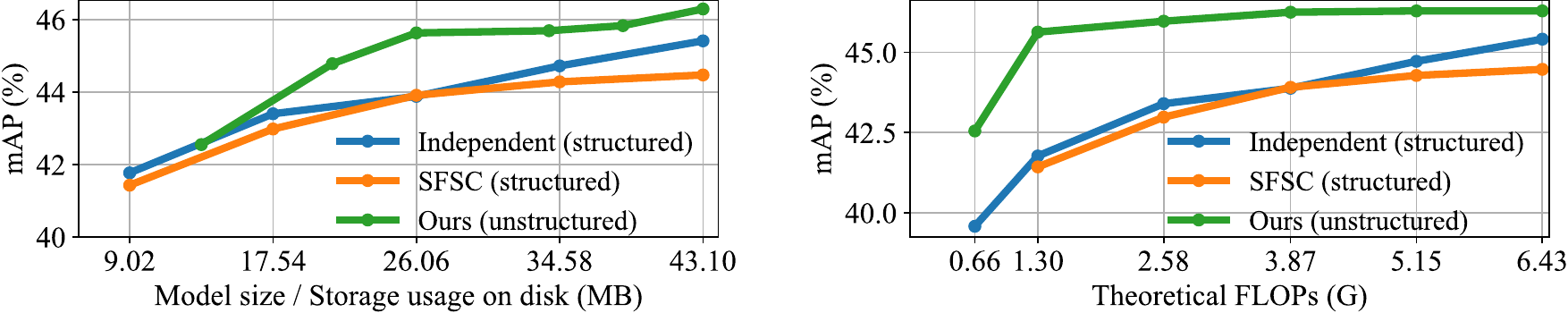}
\vspace{-1mm}
\caption{Comparison of mAP with subnetworks of different model sizes (storage usages on disk) and theoretical FLOPs.}
\label{Fig:storage_usage}
\vspace{-1mm}
\end{figure*}

\section{Additional implementation details}

\noindent\textbf{Training setup.}
We train the proposed models on two NVIDIA GeForce RTX 3090 GPUs with a batch size of 64, following the training protocols established by previous studies~\cite{advbct,inshop_SGD, fastreid} on various benchmarks. On GLDv2~\cite{gldv2}, we train Convolutional Neural Networks (CNNs), including ResNet~\cite{Resnet}, MobileNet-V2~\cite{Mobilenetv2}, and ResNeXt~\cite{Resnext}, for 30 epochs using the Stochastic Gradient Descent (SGD) optimizer with a base learning rate of 0.1, milestones at epochs [5, 10, 20], and a weight decay of 5e-4. For ViT-Small~\cite{ViT}, we use the AdamW optimizer, training for 30 epochs with a base learning rate of 3e-5 and a cosine decay scheduler with three epochs of linear warm-up. On the In-shop dataset~\cite{inshop}, we optimize ResNet-18 for 200 epochs with SGD, a base learning rate of 0.1, milestones at [50, 100, 150], and a weight decay of 5e-4. On VeRi-776~\cite{VeRi776}, ResNet-18 is trained using SGD for 60 epochs with a base learning rate of 0.01, employing a Cosine Annealing Learning Rate Scheduler after the $30$-th epoch.

\vspace{1mm}
\noindent\textbf{Adaptive BatchNorm.}
We provide a detailed explanation of Adaptive BatchNorm~\cite{Adaptive_BN}, which is employed to address the significant discrepancy in the mean and variance of Batch Normalization (BN) layers across subnetworks of different capacities. Specifically, we set the network to training mode, freeze all learnable parameters, reset the mean and variance of BN layers to zero, and perform forward propagation using a subset of the training dataset to compute the updated statistics after training. The amounts of data used for Adaptive BatchNorm are as follows: for GLDv2, 1/30 of the training dataset is utilized, while for InShop and VeRi-776, the entire training dataset is used.

\section{Pseudo algorithm}
\vspace{2mm}
We provide the algorithm description of the optimization process in Algorithm~\ref{pseudo algorithm}.

\section{More analysis and discussions}

\vspace{1mm}
\noindent\textbf{Additional analyses of hyperparameter $N$.}
We conducted additional analytical experiments to evaluate the impact of the pre-defined number of subnetworks, $N$, on model training, as illustrated in Figure~\ref{Fig:Model Number}. For $N \leq 6$, both the dense network and the subnetworks show improved performance with increasing $N$, indicating that optimizing more subnetworks jointly benefits learning more accurate rankings of the connections. However, as $N$ continues to increase, performance starts to degrade. This decline can be attributed to the increased difficulty in optimizing PrunNet, particularly due to the more intractable gradient conflicts arising from the larger number of subnetworks.

\vspace{1mm}
\noindent\textbf{Analyses of the hyperparameter $\alpha$.}
As presented in Eq. (5) in the main manuscript, we employ a hyperparameter $\alpha$ to control the influence of the conflicting degree on the weight. We conducted experiments to analyze the effect of $\alpha$. Notably, when $\alpha$ is set to 0, the method is simplified to direct gradient integration after projection. Figure~\ref{Fig:hyperparameter_alpha} illustrates the self-test and cross-test performance across different values of $\alpha$. The results indicate that the best performance is achieved at $\alpha = 0.5$. Setting $\alpha$ to a large value causes the optimization to be dominated by gradients with minimal conflict, which hinders the effective convergence of the other subnetworks and results in degraded performance. Consequently, we set $\alpha$ to 0.5 for all experiments.

\vspace{1mm}
\noindent\textbf{More visualizations.}
We visualize the cosine similarities between the gradient vectors of a single convolutional kernel in the dense network and each subnetwork, as shown in Figure~\ref{Fig:Visualize cos similarity}. We can observe that the gradient vector of each subnetwork conflicts with that of the dense network at the beginning of the training, evidenced by the negative cosine similarity. As training progresses, negative cosine similarities in our method occur only occasionally and are primarily observed in the smallest subnetwork, \ie~$\phi_{20\%}$. In contrast, the subnetworks trained with the BCT-S method encounter negative cosine similarities more frequently. This indicates that our method is more effective in alleviating gradient conflicts. Besides, we observe lower cosine similarities in the sparser subnetworks, which can be attributed to the fact that they share less weight with the dense network.

We also visualize the loss convergence curves of our method and BCT-S on GLDv2, as shown in Figure~\ref{Fig:Visualize loss scale}. At the beginning of training, the losses for both methods decline sharply. However, as training progresses, BCT-S struggles to decrease the losses of subnetworks further. The losses of subnetworks exhibit substantial inconsistency with that of the dense network. In contrast, when training PrunNet with our method, the losses of all networks remain consistent and converge to lower values.

We show additional visualization of feature distributions across the dense network and different capacities of subnetworks in Figure~\ref{Fig:feature_distribution}. All subnetworks exhibit feature distributions consistent with the dense network on Market-1501~\cite{MARKET1501} and MSMT17~\cite{MSMT17} datasets, demonstrating the effectiveness of our proposed method.

\vspace{1mm}
\noindent\textbf{Better performance than independent learning.}
In our proposed algorithm, the compatible losses ${\mathcal{L}_1, \mathcal{L}_2, \ldots, \mathcal{L}_N}$ can be interpreted as regularization terms applied to the dense network. These regularization terms are designed to encourage a small subset of weights within the network to play the role of the entire network, enabling accurate classification of input samples. Essentially, these regularization terms, along with the corresponding parameter-sharing subnetworks, promote the sparsity of PrunNet, thereby enhancing its generalization ability. Consequently, dense networks optimized using our method exhibit superior performance on various benchmarks compared to those trained independently, as demonstrated by our experimental results.

\section{Detailed experimental results}
In this section, we present the detailed experimental results over the landmark benchmarks, including RParis~\cite{rparis&roxford}, ROxford~\cite{rparis&roxford}, and GLDv2-test~\cite{gldv2}.

Table~\ref{Tab:performance_landmark_supple} reports the performance of the dense network and subnetworks at pre-determined capacities. Our method outperforms the others in terms of both self-test and cross-test performance for the dense network and most subnetworks across these three datasets.

The detailed experimental results using different architectures are shown in Table~\ref{Tab: different model architecture supplement}. Our method achieves the best performance over RParis, Roxford, and GLDv2-test on these representative architectures, indicating its strong generalization ability.

The detailed results of the experiments for simulating the deployment demand on new platforms are shown in Table~\ref{Tab:new capacity supple}. For the methods without our PrunNet, we employ BCT~\cite{BCT} or SSPL~\cite{wu2023_SSPL} to train the subnetwork at 10\% capacity compatible with the dense network, while for the methods with PrunNet, we conduct pruning by choosing the parameters with top-10\% score. Our method achieves the best performance of the subnetwork at 10\% capacity, demonstrating the effectiveness of our method and the flexibility for multi-platform deployments.

We also present detailed results of ablation studies on each landmark dataset in Table~\ref{Tab:ablation_suppl}. These detailed experimental results are consistent with the average results reported in the main manuscript, confirming the effectiveness of the proposed techniques.

\section{Experiments on additional benchmarks}
We carry out additional experiments on the following datasets to validate the generalization of our method: \textbf{(1) Market-1501~\cite{MARKET1501}}: A person re-identification dataset containing 32,668 images of 1,501 identities captured by 6 cameras. We use the standard split of 12,936 training images (751 identities) and 19,732 testing images (750 identities). \textbf{(2) MSMT17~\cite{MSMT17}}: A large-scale person re-identification dataset with 126,441 images of 4,101 identities captured by 15 cameras. We adopt the split of 32,621 training images (1,041 identities) and 93,820 testing images (3,060 identities). \textbf{(3) CUB-200-2011~\cite{CUB200}}: A fine-grained bird classification dataset with 11,788 images of 200 bird species. We use the standard split of 5,994 training images and 5,794 testing images.

The experimental results are presented in Table~\ref{Tab: market1501 results}, Table~\ref{Tab: MSMT17 results} and Table~\ref{Tab: CUB200 results}, respectively. For Market-1501 and MSMT17 experiments, we employ ResNet-18 as the backbone while adopting ViT-S for CUB-200 experiments. Our method achieves state-of-the-art performance on both self-test and cross-test, validating the effectiveness and generalization of our proposed PrunNet. In particular, we found that CUB-200 with 5,994 training images is insufficient to train ViT-S from scratch. Hence, we pretrained all models on ImageNet-1K~\cite{ImageNet} before fine-tuning them on CUB-200.

\section{Further exploration on structured pruning}

Unlike structured pruning which preserves contiguous parameter blocks compatible with hardware computation units, unstructured pruning produces irregular sparse parameters, making it challenging to achieve actual acceleration on hardware implementations. To demonstrate the practical advantages of our method implemented by unstructured pruning, we present the storage usage (in the COO format) and theoretical FLOPs in Figure~\ref{Fig:storage_usage}.

We further conduct structured pruning experiments to explore a hardware-efficient method to generate compatible subnetworks. To achieve this, we implement a kernel-level score aggregation scheme, where pruning decisions are made by averaging importance scores within each convolutional kernel and removing kernels with the lowest aggregated scores. This approach enables PrunNet to directly leverage structured pruning mechanisms while maintaining architectural integrity. As presented in Table~\ref{Tab: structured}, the structured pruning variant exhibits a moderate performance drop compared to the unstructured one, which is consistent with typical trends. Nevertheless, it outperforms SFSC, demonstrating its potential for structured sparsity. We will continue exploring structured PrunNet in future work.

\section{Convergence analyses}
In this section, we provide theoretical analyses of the convergence of our PrunNet and optimization algorithm.

\subsection{Convergence analyses of greedy pruning}
We analyze the convergence of greedy pruning in the following.
According to the gradient calculated by Eq. (2) in the main manuscript, the update of score $s_{ij}^l$ can be formulated as follows:
\begin{equation}
\label{eq:score update}
\tilde{s}_{ij}^l = s_{ij}^l - \eta \frac{\partial \mathcal{L}(\mathcal{I}_i^l)}{\partial \mathcal{I}^{l}_i} w^{l}_{ij} \mathcal{Z}^{l-1}_j.
\end{equation}
If the connection $(i,j)$ is replaced by $(i,k)$ after the update, we can conclude that $s_{ij}^l > s_{ik}^l$ but $\tilde{s}_{ij}^l<\tilde{s}_{ik}^l$. Hence we have the following inequality:
\begin{equation}
\label{eq:score update2}
\tilde{s}_{ij}^l - s_{ij}^l < \tilde{s}_{ik}^l - s_{ik}^l.
\end{equation}
Based on Eq.~\eqref{eq:score update}, we can derive the inequality:
\begin{equation}
\label{eq:score update3}
- \eta \frac{\partial \mathcal{L}(\mathcal{I}_i^l)}{\partial \mathcal{I}^{l}_i} w^{l}_{ij} \mathcal{Z}^{l-1}_j < - \eta \frac{\partial \mathcal{L}(\mathcal{I}_i^l)}{\partial \mathcal{I}^{l}_i} w^{l}_{ik} \mathcal{Z}^{l-1}_k.
\end{equation}
We denote $\tilde{\mathcal{I}}_{i}^l$ as the new input to the $i$-th neuron at the $l$-th layer $n^{l}_i$ after the replacement, and denote $\tilde{w}_{ik}^l$ as the new weight of the connection between $n^{l}_i$ and $n^{l-1}_k$. Our goal is to prove the convergence of the loss, which can be formulated as $\mathcal{L}(\tilde{\mathcal{I}}_{i}^l) < \mathcal{L}(\mathcal{I}_i^l)$.
According to Eq. (1) in the main manuscript, we have:
\begin{equation}
\label{eq:input sub}
    \mathcal{\tilde{I}}_i^l - \mathcal{I}_i^l = \tilde{w}_{ik}^l\mathcal{Z}_k^{l-1} - w_{ij}^l\mathcal{Z}_j^{l-1}.
\end{equation}
Assuming the loss is smooth and $\tilde{\mathcal{I}_i^l}$ is close to $\mathcal{I}_i^l$, we can perform a Taylor expansion of the loss at $\mathcal{I}_i^l$ ignoring the second-order term, as shown in the follows:
\begin{equation}
\label{eq:loss taylor}
\begin{split}
    & \mathcal{L}(\mathcal{\tilde{I}}_i^l) = \mathcal{L}(\mathcal{I}_i^l+(\tilde{\mathcal{I}}_i^l-\mathcal{I}_i^l))\\
    &\leq \mathcal{L}(\mathcal{I}_i^l) + \frac{\partial\mathcal{L}(\mathcal{I}_i^l)}{\partial \mathcal{I}_i^l}(\tilde{\mathcal{I}}_i^l-\mathcal{I}_i^l)\\
    & = \mathcal{L}(\mathcal{I}_i^l) + \frac{\partial\mathcal{L}(\mathcal{I}_i^l)}{\partial \mathcal{I}_i^l}(\tilde{w}_{ik}^l\mathcal{Z}_k^{l-1} - w_{ij}^l\mathcal{Z}_j^{l-1})\\
    & = \mathcal{L}(\mathcal{I}_i^l) +\frac{\partial\mathcal{L}(\mathcal{I}_i^l)}{\partial \mathcal{I}_i^l}((w_{ik}^l-\eta \frac{\partial\mathcal{L}}{\partial w_{ik}^l})\mathcal{Z}_k^{l-1} - w_{ij}^l\mathcal{Z}_j^{l-1})\\
    & = \mathcal{L}(\mathcal{I}_i^l) +\frac{\partial\mathcal{L}(\mathcal{I}_i^l)}{\partial \mathcal{I}_i^l}(w_{ik}^l\mathcal{Z}_k^{l-1} - w_{ij}^l\mathcal{Z}_j^{l-1}) \\
    &   \hspace{1em}    -\eta \frac{\partial\mathcal{L}(\mathcal{I}_i^l)}{\partial \mathcal{I}_i^l}  \frac{\partial\mathcal{L}}{\partial w_{ik}^l}\mathcal{Z}_k^{l-1}\\
    & = \mathcal{L}(\mathcal{I}_i^l) +\frac{\partial\mathcal{L}(\mathcal{I}_i^l)}{\partial \mathcal{I}_i^l}(w_{ik}^l\mathcal{Z}_k^{l-1} - w_{ij}^l\mathcal{Z}_j^{l-1}) -\eta (\frac{\partial\mathcal{L}}{\partial w_{ik}^l})^2. \\
\end{split}
\end{equation}
From Eq.~\eqref{eq:score update3}, we have $\frac{\partial\mathcal{L}(\mathcal{I}_i^l)}{\partial \mathcal{I}_i^l}(w_{ik}^l\mathcal{Z}_k^{l-1} - w_{ij}^l\mathcal{Z}_j^{l-1}) < 0$. Thus we have proven that $\mathcal{L}(\mathcal{\tilde{I}}_i^l)< \mathcal{L}(\mathcal{I}_i^l)$, indicating the convergence of our greedy pruning scheme.

\subsection{Convergence analyses of gradient integration}
\vspace{2mm}

We analyze the convergence of the proposed conflict-aware gradient integration algorithm using a two-task learning example, where two losses $\mathcal{L}_1$ and $\mathcal{L}_2$ are optimized simultaneously. In this case, the network is optimized with the total loss $\mathcal{L} = \mathcal{L}_1 + \mathcal{L}_2$ where conflict-aware gradient integration is introduced to handle the gradient conflicting issue. We assume that \(\mathcal{L}_1\) and \(\mathcal{L}_2\) are convex and differentiable, and that the gradient of \(\mathcal{L}\) is \(L\)-Lipschitz continuous with \(L > 0\). A learning rate $\eta \leq \frac{1}{L}$ is used in the conflict-aware gradient integration scheme to update the parameters. Our goal is to prove $\mathcal{L}(\tilde{\theta}) < \mathcal{L}(\theta)$, where $\theta$ is the parameters, $\tilde{\theta}$ is the new parameters updated with our conflict-aware gradient integration scheme. 

Denoting the gradients of \(\mathcal{L}_1\) and \(\mathcal{L}_2\) by \(\bm{g}_1\) and \(\bm{g}_2\), respectively, if their cosine similarity $\left \langle\bm g_1,\bm g_2\right \rangle \geq 0$, we directly calculate the summation of $\bm g_1$ and $\bm g_2$, which equals to the gradient of $\mathcal{L}$, to update the network.
Given that $\eta \leq \frac{1}{L}$, the total loss $\mathcal{L}$ will decrease unless $\nabla \mathcal{L} = 0$ in this situation. Next we discuss the situation where $\left \langle\bm g_1,\bm g_2\right \rangle < 0$. Assuming that $\nabla \mathcal{L}$ is \(L\)-Lipschitz continuous, we can conclude that $\nabla^2 \mathcal{L}(\theta) - LI$ is a negative semi-definite matrix. We then conduct a quadratic expansion of $\mathcal{L}$ around $\mathcal{L}(\theta)$, which leads to the following inequality:

\begin{equation}
\label{eq:loss_quadratic}
\begin{split}
    & \mathcal{L}(\tilde{\theta}) \leq \mathcal{L}(\theta) + \nabla \mathcal{L}(\theta)^T(\tilde{\theta} - \theta) + \frac{1}{2}\nabla^2\mathcal{L}(\theta)\parallel \tilde{\theta} - \theta \parallel^2\\
    &\hspace{2em} \leq \mathcal{L}(\theta) + \nabla \mathcal{L}(\theta)^T(\tilde{\theta} - \theta) + \frac{1}{2}L\parallel \tilde{\theta} - \theta \parallel^2. \\
\end{split}
\end{equation}
Based on Eq. (3) in the main manuscript, we have: 
\begin{equation}
\label{eq:theta_upgrade}
    \tilde{\theta} - \theta = - \eta \bm {\tilde{g}}= - n\eta(a \bm{\hat{g}}_1 + b \bm{\hat{g}}_2),
\end{equation}
where $\bm{\hat{g}}_1$ and $\bm{\hat{g}}_2$ denote the gradient after projection, $a$ and $b$ denote the cosine similarity between $(\bm g_1, \bm{\hat{g}}_1)$ and $(\bm g_2, \bm{\hat{g}}_2)$, respectively. $n$ represents the normalization coefficient, whose value equals $\frac{2}{a+b}$. Considering that $\nabla \mathcal{L}(\theta) = \bm g =\bm g_1 + \bm g_2$, Eq.~\eqref{eq:loss_quadratic} can be reformulated as: 
\begin{equation}
\label{eq:loss taylor_1}
\begin{aligned}
    & \mathcal{L}(\tilde{\theta}) \leq \mathcal{L}(\theta) - n\eta \bm g^T (a \bm {\hat{g}}_1 + b \bm {\hat{g}}_2) + \frac{1}{2}n^2 L \eta^2  \parallel a \bm {\hat{g}}_1 + b \bm {\hat{g}}_2 \parallel^2 \\ 
    & \hspace{2em}\leq \mathcal{L}(\theta) - n\eta \bm g^T (a \bm {\hat{g}}_1 + b \bm {\hat{g}}_2) + \frac{1}{2} n^2 \eta  \parallel a \bm {\hat{g}}_1 + b \bm {\hat{g}}_2 \parallel^2 \\ 
    &\hspace{2em} = \mathcal{L}(\theta) - n\eta (\bm g_1 + \bm g_2)^T(a \bm{\hat{g}}_1 + b \bm{\hat{g}}_2)\\
    & \hspace{3em}+ \frac{1}{2} n^2 \eta (a^2 \parallel \bm{\hat{g}}_1 \parallel^2  + b^2 \parallel \bm{\hat{g}}_2\parallel^2 + 2ab \bm{\hat{g}}_1 \cdot \bm{\hat{g}}_2) \\ 
    &\hspace{2em} = \mathcal{L}(\theta) - n\eta (a \bm g_1 \cdot \bm{\hat{g}}_1 + b \bm g_2 \cdot \bm{\hat{g}}_2+ a \bm{\hat{g}}_1 \cdot \bm g_2 + b \bm{\hat{g}}_2 \cdot \bm g_1 \\
    &\hspace{3em}- \frac{1}{2}na^2 \parallel \bm{\hat{g}}_1 \parallel^2 - \frac{1}{2}nb^2\parallel \bm{\hat{g}}_2\parallel^2 - nab \bm{\hat{g}}_1 \cdot \bm{\hat{g}}_2).\\
\end{aligned}
 \end{equation}
Given that $\bm{\hat{g}}_1 \cdot \bm g_2=0$ and $\bm{\hat{g}}_2 \cdot \bm g_1=0$, we can derive:
\begin{equation}
\label{eq:loss taylor_2}
\begin{aligned} 
    & \mathcal{L}(\tilde{\theta}) \leq \mathcal{L}(\theta) - n\eta (a \bm g_1 \cdot \bm{\hat{g}}_1 + b \bm g_2 \cdot \bm{\hat{g}}_2 \\
    &\hspace{3em}- \frac{1}{2}na^2 \parallel \bm{\hat{g}}_1 \parallel^2 - \frac{1}{2}nb^2\parallel \bm{\hat{g}}_2\parallel^2 - nab \bm{\hat{g}}_1 \cdot \bm{\hat{g}}_2).\\
\end{aligned}
 \end{equation}
Herein $a$ and $b$ are the cosine similarity between $(\bm g_1, \bm{\hat{g}}_1)$ and $(\bm g_2, \bm{\hat{g}}_2)$, respectively. We have 

\begin{equation}
\label{eq:loss taylor_3}
\begin{aligned} 
    & a \bm g_1 \cdot \bm{\hat g}_1 = a^2\parallel \bm g_1 \parallel \parallel \bm{\hat g}_1\parallel = a \parallel \bm{\hat g}_1\parallel^2,\\
    & b \bm g_2 \cdot \bm{\hat g}_2 = b^2\parallel \bm g_2 \parallel \parallel \bm{\hat g}_2\parallel = b \parallel \bm{\hat g}_2\parallel^2.\\
\end{aligned}
 \end{equation}
Then we get:
\begin{equation}
\label{eq:loss taylor_4}
\begin{aligned} 
    &\mathcal{L}(\tilde{\theta}) \leq \mathcal{L}(\theta) - n\eta (a\parallel \bm{\hat g}_1\parallel^2 + b\parallel \bm{\hat g}_2\parallel^2 \\
    &\hspace{3em}- \frac{1}{2}na^2 \parallel \bm{\hat{g}}_1 \parallel^2 - \frac{1}{2}nb^2\parallel \bm{\hat{g}}_2\parallel^2 - nab \bm{\hat{g}}_1 \cdot \bm{\hat{g}}_2)\\
    & \hspace{2em} = \mathcal{L}(\theta) - n\eta ((a-\frac{1}{2}na^2)\parallel \bm{\hat g}_1\parallel^2 \\
    &\hspace{3em} + ((b-\frac{1}{2}nb^2)\parallel \bm{\hat g}_2\parallel^2
     -  nab \bm{\hat{g}}_1 \cdot \bm{\hat{g}}_2)\\
     & \hspace{2em} = \mathcal{L}(\theta) - n\eta \frac{ab}{a+b}(\parallel \bm{\hat g}_1\parallel^2 + \parallel \bm{\hat g}_2\parallel^2 - 2 \bm{\hat{g}}_1 \cdot \bm{\hat{g}}_2))\\
     & \hspace{2em} = \mathcal{L}(\theta) - n\eta \frac{ab}{a+b} (\parallel \bm{\hat{g}}_1 - \bm{\hat{g}}_2 \parallel^2).
\end{aligned}
 \end{equation}

Since the angle between the vectors before and after projection is less than $\frac{\pi}{2}$, we have $a,b \in (0,1)$ and $\frac{ab}{a+b}>0$. Thus, we have proven that $\mathcal{L}(\tilde{\theta}) < \mathcal{L}(\theta)$, indicating the convergence of our conflict-aware gradient integration scheme.

{
    \small
    \bibliographystyle{ieeenat_fullname}
    \bibliography{main}

\begin{thebibliography}{55}
\providecommand{\natexlab}[1]{#1}
\providecommand{\url}[1]{\texttt{#1}}
\expandafter\ifx\csname urlstyle\endcsname\relax
  \providecommand{\doi}[1]{doi: #1}\else
  \providecommand{\doi}{doi: \begingroup \urlstyle{rm}\Url}\fi

\bibitem[Bai et~al.(2022)Bai, Jiao, Lou, Wu, Liu, Feng, and Duan]{dual-tuning}
Yan Bai, Jile Jiao, Yihang Lou, Shengsen Wu, Jun Liu, Xuetao Feng, and Ling-Yu Duan.
\newblock Dual-tuning: Joint prototype transfer and structure regularization for compatible feature learning.
\newblock \emph{IEEE Transactions on Multimedia}, pages 7287--7298, 2022.

\bibitem[Bengio et~al.(2013)Bengio, L{\'e}onard, and Courville]{Straight-through-Estimator}
Yoshua Bengio, Nicholas L{\'e}onard, and Aaron Courville.
\newblock Estimating or propagating gradients through stochastic neurons for conditional computation.
\newblock \emph{arXiv preprint arXiv:1308.3432}, pages 1--12, 2013.

\bibitem[Budnik and Avrithis(2021)]{budnik2021_AML}
Mateusz Budnik and Yannis Avrithis.
\newblock Asymmetric metric learning for knowledge transfer.
\newblock In \emph{Proceedings of the IEEE/CVF Conference on Computer Vision and Pattern Recognition}, pages 8228--8238, 2021.

\bibitem[Caruana(1997)]{multi-task-learning}
Rich Caruana.
\newblock Multitask learning.
\newblock \emph{Machine learning}, 28:\penalty0 41--75, 1997.

\bibitem[Chen et~al.(2016)Chen, Krishna, Emer, and Sze]{NPU_accelerate_2}
Yu-Hsin Chen, Tushar Krishna, Joel~S Emer, and Vivienne Sze.
\newblock Eyeriss: An energy-efficient reconfigurable accelerator for deep convolutional neural networks.
\newblock \emph{IEEE Journal of Solid-State Circuits}, 52\penalty0 (1):\penalty0 127--138, 2016.

\bibitem[Chen et~al.(2018)Chen, Badrinarayanan, Lee, and Rabinovich]{Gradnorm}
Zhao Chen, Vijay Badrinarayanan, Chen-Yu Lee, and Andrew Rabinovich.
\newblock Gradnorm: Gradient normalization for adaptive loss balancing in deep multitask networks.
\newblock In \emph{International Conference on Machine Learning}, pages 794--803. PMLR, 2018.

\bibitem[Datta et~al.(2008)Datta, Joshi, Li, and Wang]{ImageRetrieval2008}
Ritendra Datta, Dhiraj Joshi, Jia Li, and James~Z Wang.
\newblock Image retrieval: Ideas, influences, and trends of the new age.
\newblock \emph{ACM Computing Surveys (Csur)}, 40\penalty0 (2):\penalty0 1--60, 2008.

\bibitem[Deng et~al.(2009)Deng, Dong, Socher, Li, Li, and Fei-Fei]{ImageNet}
Jia Deng, Wei Dong, Richard Socher, Li-Jia Li, Kai Li, and Li Fei-Fei.
\newblock Imagenet: A large-scale hierarchical image database.
\newblock In \emph{2009 IEEE conference on computer vision and pattern recognition}, pages 248--255. Ieee, 2009.

\bibitem[D{\'e}sid{\'e}ri(2012)]{MGDA}
Jean-Antoine D{\'e}sid{\'e}ri.
\newblock Multiple-gradient descent algorithm (mgda) for multiobjective optimization.
\newblock \emph{Comptes Rendus Mathematique}, 350\penalty0 (5-6):\penalty0 313--318, 2012.

\bibitem[Diffenderfer and Kailkhura(2021)]{MPT}
James Diffenderfer and Bhavya Kailkhura.
\newblock Multi-prize lottery ticket hypothesis: Finding accurate binary neural networks by pruning a randomly weighted network.
\newblock \emph{arXiv preprint arXiv:2103.09377}, pages 1--23, 2021.

\bibitem[Dosovitskiy(2020)]{ViT}
Alexey Dosovitskiy.
\newblock An image is worth 16x16 words: Transformers for image recognition at scale.
\newblock \emph{arXiv preprint arXiv:2010.11929}, pages 1--22, 2020.

\bibitem[Duggal et~al.(2021)Duggal, Zhou, Yang, Xiong, Xia, Tu, and Soatto]{duggal2021_CMP_NAS}
Rahul Duggal, Hao Zhou, Shuo Yang, Yuanjun Xiong, Wei Xia, Zhuowen Tu, and Stefano Soatto.
\newblock Compatibility-aware heterogeneous visual search.
\newblock In \emph{Proceedings of the IEEE/CVF Conference on Computer Vision and Pattern Recognition}, pages 10723--10732, 2021.

\bibitem[Frankle and Carbin(2018)]{LTH}
Jonathan Frankle and Michael Carbin.
\newblock The lottery ticket hypothesis: Finding sparse, trainable neural networks.
\newblock \emph{arXiv preprint arXiv:1803.03635}, pages 1--42, 2018.

\bibitem[He et~al.(2016)He, Zhang, Ren, and Sun]{Resnet}
Kaiming He, Xiangyu Zhang, Shaoqing Ren, and Jian Sun.
\newblock Deep residual learning for image recognition.
\newblock In \emph{Proceedings of the IEEE Conference on Computer Vision and Pattern Recognition}, pages 770--778, 2016.

\bibitem[He et~al.(2023)He, Liao, Liu, Liu, Cheng, and Mei]{fastreid}
Lingxiao He, Xingyu Liao, Wu Liu, Xinchen Liu, Peng Cheng, and Tao Mei.
\newblock Fastreid: A pytorch toolbox for general instance re-identification.
\newblock In \emph{Proceedings of the 31st ACM International Conference on Multimedia}, pages 9664--9667, 2023.

\bibitem[Kang et~al.(2022)Kang, Mina, Madjid, Yoon, Hasegawa-Johnson, Hwang, and Yoo]{WSN}
Haeyong Kang, Rusty John~Lloyd Mina, Sultan Rizky~Hikmawan Madjid, Jaehong Yoon, Mark Hasegawa-Johnson, Sung~Ju Hwang, and Chang~D Yoo.
\newblock Forget-free continual learning with winning subnetworks.
\newblock In \emph{International Conference on Machine Learning}, pages 10734--10750. PMLR, 2022.

\bibitem[Kendall et~al.(2018)Kendall, Gal, and Cipolla]{MTL_uncertainty}
Alex Kendall, Yarin Gal, and Roberto Cipolla.
\newblock Multi-task learning using uncertainty to weigh losses for scene geometry and semantics.
\newblock In \emph{Proceedings of the IEEE Conference on Computer Vision and Pattern Recognition}, pages 7482--7491, 2018.

\bibitem[Krizhevsky et~al.(2009)Krizhevsky, Hinton, et~al.]{CIFAR}
Alex Krizhevsky, Geoffrey Hinton, et~al.
\newblock Learning multiple layers of features from tiny images.
\newblock pages 1--60, 2009.

\bibitem[Li et~al.(2020)Li, Wu, Su, and Wang]{Adaptive_BN}
Bailin Li, Bowen Wu, Jiang Su, and Guangrun Wang.
\newblock Eagleeye: Fast sub-net evaluation for efficient neural network pruning.
\newblock In \emph{Computer Vision--ECCV 2020: 16th European Conference, Glasgow, UK, August 23--28, 2020, Proceedings, Part II 16}, pages 639--654. Springer, 2020.

\bibitem[Liu et~al.(2021)Liu, Liu, Jin, Stone, and Liu]{CAGrad}
Bo Liu, Xingchao Liu, Xiaojie Jin, Peter Stone, and Qiang Liu.
\newblock Conflict-averse gradient descent for multi-task learning.
\newblock \emph{Advances in Neural Information Processing Systems}, 34:\penalty0 18878--18890, 2021.

\bibitem[Liu et~al.(2016{\natexlab{a}})Liu, Liu, Mei, and Ma]{VeRi776}
Xinchen Liu, Wu Liu, Tao Mei, and Huadong Ma.
\newblock A deep learning-based approach to progressive vehicle re-identification for urban surveillance.
\newblock In \emph{Computer Vision--ECCV 2016: 14th European Conference, Amsterdam, The Netherlands, October 11-14, 2016, Proceedings, Part II 14}, pages 869--884. Springer, 2016{\natexlab{a}}.

\bibitem[Liu et~al.(2016{\natexlab{b}})Liu, Luo, Qiu, Wang, and Tang]{inshop}
Ziwei Liu, Ping Luo, Shi Qiu, Xiaogang Wang, and Xiaoou Tang.
\newblock Deepfashion: Powering robust clothes recognition and retrieval with rich annotations.
\newblock In \emph{Proceedings of the IEEE Conference on Computer Vision and Pattern Recognition}, pages 1096--1104, 2016{\natexlab{b}}.

\bibitem[Meng et~al.(2021)Meng, Zhang, Xu, and Zhou]{LCE}
Qiang Meng, Chixiang Zhang, Xiaoqiang Xu, and Feng Zhou.
\newblock Learning compatible embeddings.
\newblock In \emph{Proceedings of the IEEE/CVF International Conference on Computer Vision}, pages 9939--9948, 2021.

\bibitem[Mu{\~n}oz-Mart{\'\i}nez et~al.(2023)Mu{\~n}oz-Mart{\'\i}nez, Garg, Pellauer, Abell{\'a}n, Acacio, and Krishna]{Flexagon}
Francisco Mu{\~n}oz-Mart{\'\i}nez, Raveesh Garg, Michael Pellauer, Jos{\'e}~L Abell{\'a}n, Manuel~E Acacio, and Tushar Krishna.
\newblock Flexagon: A multi-dataflow sparse-sparse matrix multiplication accelerator for efficient dnn processing.
\newblock In \emph{Proceedings of the 28th ACM International Conference on Architectural Support for Programming Languages and Operating Systems, Volume 3}, pages 252--265, 2023.

\bibitem[Pan et~al.(2023)Pan, Xu, Yang, He, Jiang, Guo, Qian, Zhang, Cheng, Yang, et~al.]{advbct}
Tan Pan, Furong Xu, Xudong Yang, Sifeng He, Chen Jiang, Qingpei Guo, Feng Qian, Xiaobo Zhang, Yuan Cheng, Lei Yang, et~al.
\newblock Boundary-aware backward-compatible representation via adversarial learning in image retrieval.
\newblock In \emph{Proceedings of the IEEE/CVF Conference on Computer Vision and Pattern Recognition}, pages 15201--15210, 2023.

\bibitem[Radenovi{\'c} et~al.(2018)Radenovi{\'c}, Iscen, Tolias, Avrithis, and Chum]{rparis&roxford}
Filip Radenovi{\'c}, Ahmet Iscen, Giorgos Tolias, Yannis Avrithis, and Ond{\v{r}}ej Chum.
\newblock Revisiting oxford and paris: Large-scale image retrieval benchmarking.
\newblock In \emph{Proceedings of the IEEE Conference on Computer Vision and Pattern Recognition}, pages 5706--5715, 2018.

\bibitem[Ramanujan et~al.(2020)Ramanujan, Wortsman, Kembhavi, Farhadi, and Rastegari]{What's_hidden}
Vivek Ramanujan, Mitchell Wortsman, Aniruddha Kembhavi, Ali Farhadi, and Mohammad Rastegari.
\newblock What's hidden in a randomly weighted neural network?
\newblock In \emph{Proceedings of the IEEE/CVF Conference on Computer Vision and Pattern Recognition}, pages 11893--11902, 2020.

\bibitem[Sandler et~al.(2018)Sandler, Howard, Zhu, Zhmoginov, and Chen]{Mobilenetv2}
Mark Sandler, Andrew Howard, Menglong Zhu, Andrey Zhmoginov, and Liang-Chieh Chen.
\newblock Mobilenetv2: Inverted residuals and linear bottlenecks.
\newblock In \emph{Proceedings of the IEEE Conference on Computer Vision and Pattern Recognition}, pages 4510--4520, 2018.

\bibitem[Sener and Koltun(2018)]{Multi_pareto}
Ozan Sener and Vladlen Koltun.
\newblock Multi-task learning as multi-objective optimization.
\newblock \emph{Advances in Neural Information Processing Systems}, 31:\penalty0 525--536, 2018.

\bibitem[Seo et~al.(2023)Seo, Uzunbas, Han, Cao, Zhang, Tian, and Lim]{Online-Backfilling}
Seonguk Seo, Mustafa~Gokhan Uzunbas, Bohyung Han, Sara Cao, Joena Zhang, Taipeng Tian, and Ser-Nam Lim.
\newblock Online backfilling with no regret for large-scale image retrieval.
\newblock \emph{arXiv preprint arXiv:2301.03767}, pages 1--10, 2023.

\bibitem[Shen et~al.(2020)Shen, Xiong, Xia, and Soatto]{BCT}
Yantao Shen, Yuanjun Xiong, Wei Xia, and Stefano Soatto.
\newblock Towards backward-compatible representation learning.
\newblock In \emph{Proceedings of the IEEE/CVF Conference on Computer Vision and Pattern Recognition}, pages 6368--6377, 2020.

\bibitem[Shoshan et~al.(2024)Shoshan, Linial, Bhonker, Hirsch, Zamir, Kviatkovsky, and Medioni]{shoshan2024_ensemble}
Alon Shoshan, Ori Linial, Nadav Bhonker, Elad Hirsch, Lior Zamir, Igor Kviatkovsky, and G{\'e}rard Medioni.
\newblock Asymmetric image retrieval with cross model compatible ensembles.
\newblock In \emph{Proceedings of the IEEE/CVF Winter Conference on Applications of Computer Vision}, pages 1--11, 2024.

\bibitem[Suma and Tolias(2023)]{suma2023_RAML}
Pavel Suma and Giorgos Tolias.
\newblock Large-to-small image resolution asymmetry in deep metric learning.
\newblock In \emph{Proceedings of the IEEE/CVF Winter Conference on Applications of Computer Vision}, pages 1451--1460, 2023.

\bibitem[Suo et~al.(2024)Suo, Ma, Zhu, and Yang]{ImageRetrieval2024}
Yucheng Suo, Fan Ma, Linchao Zhu, and Yi Yang.
\newblock Knowledge-enhanced dual-stream zero-shot composed image retrieval.
\newblock In \emph{Proceedings of the IEEE/CVF Conference on Computer Vision and Pattern Recognition}, pages 26951--26962, 2024.

\bibitem[Wah et~al.(2011)Wah, Branson, Welinder, Perona, and Belongie]{CUB200}
Catherine Wah, Steve Branson, Peter Welinder, Pietro Perona, and Serge Belongie.
\newblock The caltech-ucsd birds-200-2011 dataset.
\newblock 2011.

\bibitem[Wang et~al.(2023)Wang, Du, Yang, Zhang, Wang, Zhang, Yang, Huang, and Han]{ImageRetrieval2023}
Xiyue Wang, Yuexi Du, Sen Yang, Jun Zhang, Minghui Wang, Jing Zhang, Wei Yang, Junzhou Huang, and Xiao Han.
\newblock Retccl: Clustering-guided contrastive learning for whole-slide image retrieval.
\newblock \emph{Medical Image Analysis}, 83:\penalty0 102645--102645, 2023.

\bibitem[Wang(2020)]{Sparsert}
Ziheng Wang.
\newblock Sparsert: Accelerating unstructured sparsity on gpus for deep learning inference.
\newblock In \emph{Proceedings of the ACM International Conference on Parallel Architectures and Compilation Techniques}, pages 31--42, 2020.

\bibitem[Wang et~al.(2020)Wang, Tsvetkov, Firat, and Cao]{Grad_Vaccine}
Zirui Wang, Yulia Tsvetkov, Orhan Firat, and Yuan Cao.
\newblock Gradient vaccine: Investigating and improving multi-task optimization in massively multilingual models.
\newblock \emph{arXiv preprint arXiv:2010.05874}, pages 1--22, 2020.

\bibitem[Wei et~al.(2018)Wei, Zhang, Gao, and Tian]{MSMT17}
Longhui Wei, Shiliang Zhang, Wen Gao, and Qi Tian.
\newblock Person transfer gan to bridge domain gap for person re-identification.
\newblock In \emph{Proceedings of the IEEE conference on computer vision and pattern recognition}, pages 79--88, 2018.

\bibitem[Weyand et~al.(2020)Weyand, Araujo, Cao, and Sim]{gldv2}
Tobias Weyand, Andre Araujo, Bingyi Cao, and Jack Sim.
\newblock Google landmarks dataset v2-a large-scale benchmark for instance-level recognition and retrieval.
\newblock In \emph{Proceedings of the IEEE/CVF Conference on Computer Vision and Pattern Recognition}, pages 2575--2584, 2020.

\bibitem[Wu et~al.(2022)Wu, Wang, Zhou, Li, and Tian]{wu2022_CSD}
Hui Wu, Min Wang, Wengang Zhou, Houqiang Li, and Qi Tian.
\newblock Contextual similarity distillation for asymmetric image retrieval.
\newblock In \emph{Proceedings of the IEEE/CVF Conference on Computer Vision and Pattern Recognition}, pages 9489--9498, 2022.

\bibitem[Wu et~al.(2023{\natexlab{a}})Wu, Wang, Zhou, and Li]{wu2023_ROP}
Hui Wu, Min Wang, Wengang Zhou, and Houqiang Li.
\newblock A general rank preserving framework for asymmetric image retrieval.
\newblock In \emph{The Eleventh International Conference on Learning Representations}, pages 1--20, 2023{\natexlab{a}}.

\bibitem[Wu et~al.(2023{\natexlab{b}})Wu, Wang, Zhou, and Li]{wu2023_SSPL}
Hui Wu, Min Wang, Wengang Zhou, and Houqiang Li.
\newblock Structure similarity preservation learning for asymmetric image retrieval.
\newblock \emph{IEEE Transactions on Multimedia}, pages 4693--4705, 2023{\natexlab{b}}.

\bibitem[Wu et~al.(2023{\natexlab{c}})Wu, Wang, Zhou, Lu, and Li]{wu2023_AFF}
Hui Wu, Min Wang, Wengang Zhou, Zhenbo Lu, and Houqiang Li.
\newblock Asymmetric feature fusion for image retrieval.
\newblock In \emph{Proceedings of the IEEE/CVF Conference on Computer Vision and Pattern Recognition}, pages 11082--11092, 2023{\natexlab{c}}.

\bibitem[Wu et~al.(2023{\natexlab{d}})Wu, Bai, Lou, Linghu, He, and Duan]{wu2023_SFSC}
Shengsen Wu, Yan Bai, Yihang Lou, Xiongkun Linghu, Jianzhong He, and Ling-Yu Duan.
\newblock Switchable representation learning framework with self-compatibility.
\newblock In \emph{Proceedings of the IEEE/CVF Conference on Computer Vision and Pattern Recognition}, pages 15943--15953, 2023{\natexlab{d}}.

\bibitem[Xie et~al.(2017)Xie, Girshick, Doll{\'a}r, Tu, and He]{Resnext}
Saining Xie, Ross Girshick, Piotr Doll{\'a}r, Zhuowen Tu, and Kaiming He.
\newblock Aggregated residual transformations for deep neural networks.
\newblock In \emph{Proceedings of the IEEE Conference on Computer Vision and Pattern Recognition}, pages 1492--1500, 2017.

\bibitem[Xie et~al.(2023)Xie, Zhang, Xu, Zhu, and He]{xie2023_CDD}
Yi Xie, Huaidong Zhang, Xuemiao Xu, Jianqing Zhu, and Shengfeng He.
\newblock Towards a smaller student: Capacity dynamic distillation for efficient image retrieval.
\newblock In \emph{2023 IEEE/CVF Conference on Computer Vision and Pattern Recognition}, pages 16006--16015. IEEE, 2023.

\bibitem[Xie et~al.(2024)Xie, Lin, Cai, Xu, Zhang, Du, and He]{xie2024_d3still}
Yi Xie, Yihong Lin, Wenjie Cai, Xuemiao Xu, Huaidong Zhang, Yong Du, and Shengfeng He.
\newblock D3still: Decoupled differential distillation for asymmetric image retrieval.
\newblock In \emph{Proceedings of the IEEE/CVF Conference on Computer Vision and Pattern Recognition}, pages 17181--17190, 2024.

\bibitem[You et~al.(2022)You, Li, Sun, Ouyang, and Lin]{Supertickets}
Haoran You, Baopu Li, Zhanyi Sun, Xu Ouyang, and Yingyan Lin.
\newblock Supertickets: Drawing task-agnostic lottery tickets from supernets via jointly architecture searching and parameter pruning.
\newblock In \emph{European Conference on Computer Vision}, pages 674--690. Springer, 2022.

\bibitem[Yu et~al.(2020)Yu, Kumar, Gupta, Levine, Hausman, and Finn]{PCGrad}
Tianhe Yu, Saurabh Kumar, Abhishek Gupta, Sergey Levine, Karol Hausman, and Chelsea Finn.
\newblock Gradient surgery for multi-task learning.
\newblock \emph{Advances in Neural Information Processing Systems}, 33:\penalty0 5824--5836, 2020.

\bibitem[Zhai and Wu(2018)]{inshop_SGD}
Andrew Zhai and Hao-Yu Wu.
\newblock Classification is a strong baseline for deep metric learning.
\newblock \emph{arXiv preprint arXiv:1811.12649}, pages 1--12, 2018.

\bibitem[Zhang et~al.(2021)Zhang, Ge, Shen, Li, Yuan, Xu, Wang, and Shan]{hot-refresh}
Binjie Zhang, Yixiao Ge, Yantao Shen, Yu Li, Chun Yuan, Xuyuan Xu, Yexin Wang, and Ying Shan.
\newblock Hot-refresh model upgrades with regression-free compatible training in image retrieval.
\newblock In \emph{International Conference on Learning Representations}, pages 1--20, 2021.

\bibitem[Zheng et~al.(2015)Zheng, Shen, Tian, Wang, Wang, and Tian]{MARKET1501}
Liang Zheng, Liyue Shen, Lu Tian, Shengjin Wang, Jingdong Wang, and Qi Tian.
\newblock Scalable person re-identification: A benchmark.
\newblock In \emph{Proceedings of the IEEE international conference on computer vision}, pages 1116--1124, 2015.

\bibitem[Zhou et~al.(2025)Zhou, Sun, Pei, Li, and Wang]{Prototype_Perturbation}
Zikun Zhou, Yushuai Sun, Wenjie Pei, Xin Li, and Yaowei Wang.
\newblock Prototype perturbation for relaxing alignment constraints in backward-compatible learning.
\newblock \emph{arXiv preprint arXiv:2503.14824}, pages 1--11, 2025.

\bibitem[Zhu et~al.(2020)Zhu, Huang, Yang, Zhu, Zhang, and Shen]{FPGA_accelerator}
Chaoyang Zhu, Kejie Huang, Shuyuan Yang, Ziqi Zhu, Hejia Zhang, and Haibin Shen.
\newblock An efficient hardware accelerator for structured sparse convolutional neural networks on fpgas.
\newblock \emph{IEEE Transactions on Very Large Scale Integration (VLSI) Systems}, 28\penalty0 (9):\penalty0 1953--1965, 2020.

\end{thebibliography}
}

\end{document}